\documentclass{article}

\PassOptionsToPackage{numbers, compress}{natbib}

\usepackage[main,final]{neurips_2026}

\usepackage[utf8]{inputenc} %
\usepackage[T1]{fontenc}    %
\usepackage{hyperref}       %
\usepackage{url}            %
\usepackage{booktabs}       %
\usepackage{amsfonts}       %
\usepackage{nicefrac}       %
\usepackage{microtype}      %
\usepackage{xcolor}         %
\usepackage{graphicx}
\usepackage{multirow}
\usepackage{arydshln}
\usepackage{adjustbox}
\usepackage{amssymb}
\usepackage{listings}
\usepackage{xcolor}
\definecolor{figyellow}{RGB}{210,180,0}

\usepackage[most]{tcolorbox}
\usepackage{xcolor}
\usepackage{wrapfig}
\usepackage[table]{xcolor}
\usepackage{placeins}

\newcommand{\upb}{\,\ensuremath{\uparrow}}
\newcommand{\downb}{\,\ensuremath{\downarrow}}

\newtcolorbox{promptbox}[1]{
    title=#1,
    colback=blue!5!white,       %
    colframe=blue!75!black,     %
    coltitle=white,             %
    breakable,
    fontupper=\small
}

\title{Enhancing Gaze Reasoning in Vision Foundation Models for Gaze Following}

\author{%
  \textbf{Shijing Wang}\textsuperscript{1} \quad
  \textbf{Yaping Huang}\textsuperscript{1}\thanks{Corresponding author.} \quad
  \textbf{Chaoqun Cui}\textsuperscript{4} \quad
  \textbf{David Wong}\textsuperscript{5} \quad
  \textbf{Yihua Cheng}\textsuperscript{2, 3}\footnotemark[1]
  \\
  \textbf{Alexandros Neophytou}\textsuperscript{5} \quad
  \textbf{Hyung Jin Chang}\textsuperscript{2, 5}
  \\[0.5ex]
  \textsuperscript{1}Beijing Key Laboratory of Traffic Data Mining and Embodied Intelligence,\\
  Beijing Jiaotong University\\
  \textsuperscript{2}University of Birmingham \quad
  \textsuperscript{3}Beijing Institute of Technology\\
  \textsuperscript{4}Institute of Automation, Chinese Academy of Sciences \quad
  \textsuperscript{5}Microsoft, UK
}

\begin{document}
\raggedbottom

\maketitle

\begin{abstract}

Gaze following requires both scene understanding and gaze reasoning to localize the gaze target of an in-scene person. Recently, vision foundation models (VFMs) have demonstrated strong performance on this task, enabling simpler architectures while outperforming prior methods. However, we observe a key limitation of VFM-based approaches: while VFMs substantially improve scene understanding, they contribute little to gaze reasoning. As a result, existing methods often rely on semantically salient objects rather than true gaze cues, leading to degraded performance when targets are not salient.
To address this, we propose a novel training mechanism to enhance gaze reasoning in VFMs for gaze following. Our method includes: (1) a head-conditioned local LoRA, which enables localized adaptation to preserve scene token learning while improving head token learning for gaze reasoning; and (2) an out-of-cone penalty, which injects gaze cues into head tokens while aligning them with scene tokens.
Experiments on the GazeFollow and VAT datasets demonstrate that our method achieves state-of-the-art performance, with particularly strong improvements when gaze targets are not semantically salient. Our findings offer valuable insights for advancing future gaze following research.
Code will be available at \url{https://github.com/shijing-wang/HCloRA}.

\end{abstract}

\section{Introduction}

Human gaze behavior is a fundamental component of non-verbal communication~\cite{capozzi2019tracking, tafasca2023ai4autism}. The ability to follow a person's gaze is essential for understanding human behavior, and has broad applications in human-robot interaction~\cite{admoni2017social,quesada2025integrated}, assistive systems~\cite{li2022appearance,zhang2025mindeye}, and behavior understanding~\cite{kim2021assessing,hessels2025gaze}.

Gaze following is challenging because it requires both scene understanding and gaze reasoning.
The model must understand the surrounding scene to capture contextual information, while also using gaze cues to reason where the target person is looking.
Vision foundation models (VFMs) are well suited to the scene-understanding aspect of this task, as their pre-trained representations provide strong semantic and contextual priors for interpreting complex visual environments~\cite{oquab2023dinov2,bachmann2022multimae,tu2022end}.
These properties make VFMs attractive backbones for gaze following, contributing to the remarkable progress and new state-of-the-art performance achieved in this task.

\begin{figure*}[t]
    \centering
    \includegraphics[width=\linewidth]{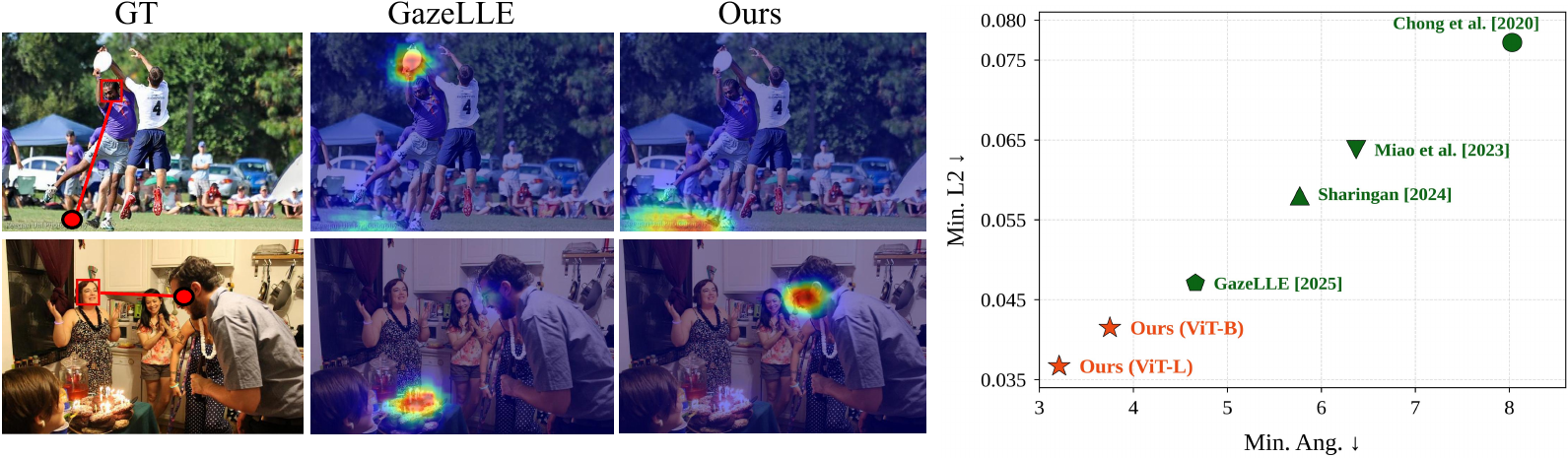}
    \caption{
    \textbf{Left}: Given an input image and the query head marked by the red box, existing VFM-based methods such as GazeLLE often predict gaze targets around semantically salient objects or interaction centers, even when they are not the true gaze target.
    In contrast, our method better localizes the gaze target by relying on true gaze cues rather than semantic saliency.
    \textbf{Right}: Comparison on the GazeFollow benchmark, where lower Min. L2 and Min. Ang. indicate better performance.
    Our method achieves state-of-the-art results, demonstrating the effectiveness of enhancing gaze reasoning in VFM-based gaze following.}
    \label{fig:first}
\end{figure*}

However, in this paper, we observe that the strong performance of VFM-based gaze following does not necessarily imply stronger gaze reasoning.
VFMs provide rich scene-level representations and are highly effective at capturing semantic and contextual regularities in visual scenes.
As shown in Fig.~\ref{fig:first} (left), existing methods can often make plausible predictions by exploiting scene priors, such as focusing on visually prominent objects, interaction centers, or semantically meaningful regions.
While these priors are helpful in many common cases, they become unreliable when the true gaze target is not the most salient region in the scene.
Consequently, current methods tend to rely on semantically salient objects rather than true gaze cues, assigning high confidence to visually plausible but incorrect regions and thus degrading localization performance.

To address this issue, we propose a novel training mechanism for VFM-based gaze following that enhances gaze reasoning without redesigning the main backbone-decoder pipeline.
Our method consists of two complementary components.
First, we introduce a head-conditioned local LoRA, which uses the queried head position to modulate low-rank residual updates inside the VFM.
This enables localized adaptation of head-relevant features while preserving the backbone's scene-level representations.
Second, we propose an out-of-cone penalty, which regularizes auxiliary gaze-evidence maps from adapted layers by penalizing probability mass outside the geometrically plausible gaze cone.
Instead of imposing a rigid gaze-cone prior, this penalty flexibly injects gaze cues into head features.
Together, these designs encourage the model to rely on true gaze cues rather than insufficient semantic saliency.

Experiments on the GazeFollow and VAT datasets demonstrate that our method achieves state-of-the-art performance, with representative results shown in Fig.~\ref{fig:first} (right).
The gains are especially significant when gaze targets are not semantically salient, showing that our method improves gaze following by strengthening gaze reasoning rather than relying on scene understanding alone.
These findings highlight the importance of explicit gaze reasoning for robust VFM-based gaze following.

In summary, our main contributions are as follows:
\begin{itemize}
    \item We identify a key limitation of current VFM-based gaze following methods: although VFMs substantially improve scene understanding, they contribute little to gaze reasoning, causing models to mainly rely on insufficient semantically salient objects rather than true gaze cues.

    \item We propose a novel training mechanism for VFM-based gaze following without redesigning the main backbone-decoder pipeline, including a head-conditioned local LoRA and an out-of-cone penalty to enhance gaze reasoning.

    \item We achieve state-of-the-art performance on the GazeFollow and VAT datasets, with strong gains in more challenging scenarios where gaze targets are not semantically salient.
\end{itemize}

\section{Related Work}

\subsection{Gaze Following}

Gaze following aims to localize the gaze point of the target person in a scene.
Recasens et al.~\cite{recasens2015they} first formulated this task by combining head information with scene context to predict a gaze heatmap.
Following this paradigm, later methods improve gaze prediction by incorporating additional cues, including depth~\cite{fang2021dual,gupta2022modular}, body pose~\cite{gupta2022modular}, 3D head orientation~\cite{horanyi2023they}, and temporal context~\cite{chong2020detecting,gupta2024exploring}.
More recently, transformer-based models have been introduced to better capture interactions between the queried person and the surrounding scene~\cite{tafasca2024sharingan,ryan2025gaze}.
These methods improve contextual modeling and highlight the importance of jointly modeling head information and scene context.

\subsection{Vision Foundation Models for Gaze Following}

Vision foundation models, such as DINOv2~\cite{oquab2023dinov2} and MultiMAE~\cite{bachmann2022multimae}, learn transferable visual representations from large-scale pretraining.
Their strong semantic and contextual understanding makes them effective backbones for gaze following. %
Recent gaze-following methods have started to build upon such pre-trained visual representations.
Sharingan~\citep{tafasca2024sharingan} represents the image as scene tokens and fuses them with person-specific gaze tokens through transformer attention.
GazeLLE~\cite{ryan2025gaze} further simplifies the architecture by using a frozen VFM backbone for scene feature extraction and injecting the queried head information into a lightweight decoder, achieving stronger performance.

\section{Preliminary}
\label{sec:analysis}

VFMs have brought performance breakthroughs to various vision tasks, including gaze following.
In this work, we aim to further improve the performance of VFM-based gaze following models.

\subsection{Observation: Semantic Shortcut on Gaze Following}

\begin{figure}[t]
    \centering
    \includegraphics[width=\linewidth]{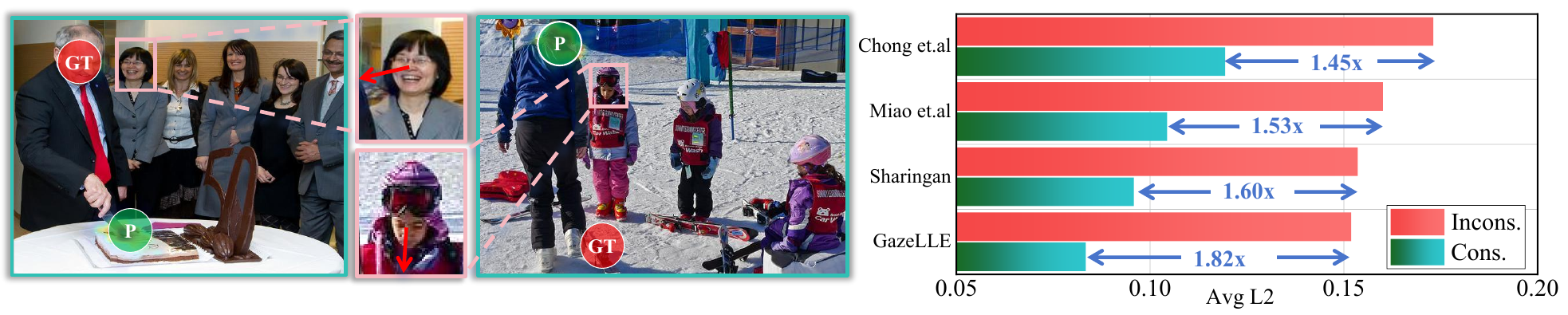}
    \caption{
    \textbf{Left:} Failure cases of GazeLLE~\citep{ryan2025gaze}, where the predicted gaze target is attracted to semantically salient regions instead of the ground-truth gaze target.
    \textbf{Right:} Average L2 error on the GazeFollow consistent and inconsistent subsets across different methods.
    All methods show larger errors on the inconsistent subset, suggesting that current gaze-following models are still affected by semantic saliency and do not always perform robust gaze reasoning.
    }
    \label{fig:analysis_examples}
\end{figure}

We start by analyzing the failure cases of recent models.
Specifically, we select  GazeLLE~\citep{ryan2025gaze}, a state-of-the-art VFM-based method, as our baseline. We sort test samples by prediction error, and visualize representative failure samples.
As shown in Fig.~\ref{fig:analysis_examples}(a), large-error predictions usually share a common pattern:
\textit{the prediction concentrates on semantically salient regions, such as manipulated objects or interaction centers, even when these regions lie in the opposite direction of the human gaze.} Please see Appendix~\ref{app:semantic_shortcut_failures} for more failure cases.

This observation indicates that such models cannot capture reliable gaze cues. Instead, they tend to infer gaze targets from scene-level semantic context, which can lead to an overestimation of the model's gaze reasoning ability when the gaze target coincides with a semantically salient object.
We refer to this behavior as a \emph{semantic shortcut}, where the model relies on semantic plausibility as a proxy for gaze following rather than performing gaze reasoning. However, we argue that a robust gaze following method should possess strong and reliable gaze reasoning capability.

\subsection{Quantitative Analysis: Performance Gap from Semantic Shortcut}
\label{sec:semantic_partition}

To further validate our hypothesis, we conduct a quantitative analysis based on the results of GazeLLE. Specifically, we aim to split the test dataset into two subsets: \textit{\textbf{consistent}}, where the gaze target is consistent with a semantically salient object, and \textit{\textbf{inconsistent}}, where it is not. If we observe a performance gap between these two subsets, it supports our hypothesis.

However, it is non-trivial to precisely define and identify semantically salient objects.
Therefore, we adopt a \textit{heuristic strategy} for this analysis. Specifically, we use Qwen3-VL-32B-Instruct~\citep{bai2025qwen3} as a semantic prober to identify contextually plausible gaze regions for one person.
Importantly, the prober is not used to predict the gaze target directly.
Instead, it is prompted to localize regions that are semantically likely to attract the target person's attention based on scene context.
Each predicted region is represented by a bounding box.
We then measure the overlap between the ground-truth gaze points and the predicted boxes. If more than half of the gaze points fall within the boxes, the sample is assigned to the \emph{consistent} subset; otherwise, it is assigned to the \emph{inconsistent} subset. For single annotation, the threshold is set to 1.
Please see Appendices~\ref{app:semantic_partition_details} and~\ref{app:semantic_partition_examples} for more details.

The results are shown in Fig.~\ref{fig:analysis_examples}(b). GazeLLE exhibits a significant performance gap between the two subsets, which validates our hypothesis. We further evaluate other gaze following methods on these subsets and observe a similar trend: all methods show a noticeable performance gap. This finding suggests that the semantic shortcut is a common issue across gaze following approaches, \textit{highlighting the importance of enhancing gaze reasoning in current methods to alleviate this problem.}

\section{Methodology}
\label{sec:method}

We propose a novel training mechanism to enhance gaze reasoning in VFMs without redesigning the main backbone-decoder pipeline. Our contributions include a head-conditioned local LoRA for head feature extraction and an out-of-cone penalty for injecting gaze cues into head features.

\subsection{Head-Conditioned Local LoRA}

Previous gaze following methods typically freeze the VFM for image tokenization~\cite{ryan2025gaze}, while using a learnable decoder for adaptation. This raises an intuitive question: \textit{Can simply fine-tuning the VFM on gaze following datasets improve gaze reasoning?}

\begin{wrapfigure}[12]{r}{0.68\linewidth}
\vspace{\dimexpr-\intextsep-4pt\relax}
\centering
\includegraphics[width=\linewidth]{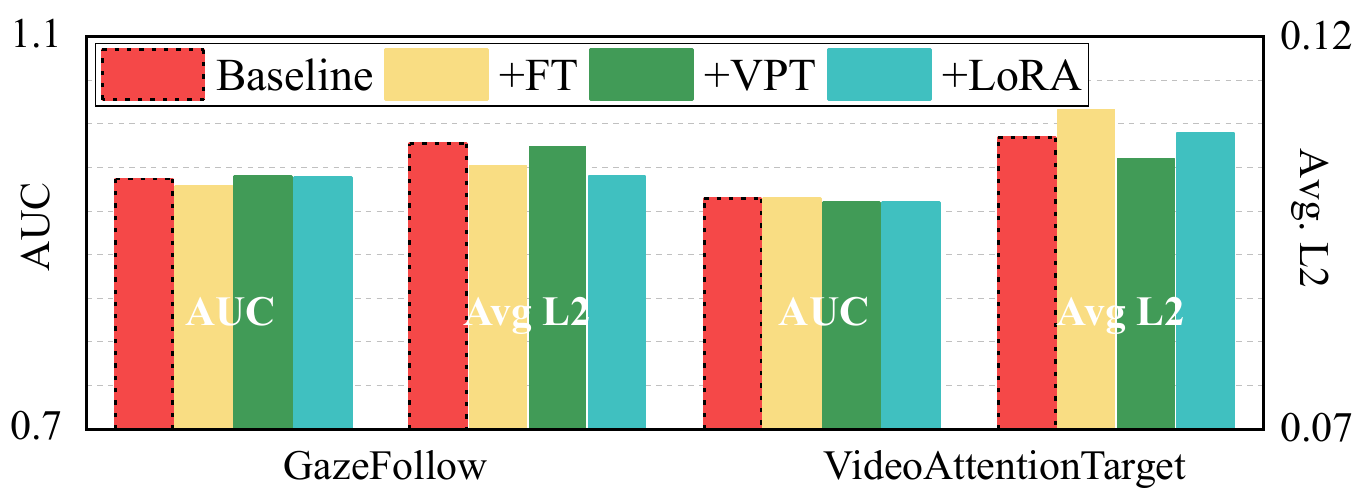}
\caption{\small We fine-tune GazeLLE on two datasets using several SoTA fine-tuning methods. Naive fine-tuning fails to consistently improve performance.}
\label{fig:naive_finetuning}
\end{wrapfigure}

\textbf{Naive fine-tuning remains insufficient.} We conduct experiments to evaluate the effectiveness of fine-tuning. As shown in Fig.~\ref{fig:naive_finetuning}, we evaluate full fine-tuning (FT), LoRA~\cite{hu2021lora}, and visual prompt tuning (VPT). Interestingly, these methods fail to deliver consistent performance improvements and even degrade performance in some cases.

We attribute this to the following insight: VFMs have strong scene understanding capabilities through large-scale pretraining on diverse image data. Fine-tuning on relatively small datasets may degrade this capability. At the same time, VFMs show limited gaze reasoning ability, meaning they struggle to precisely capture human facial and gaze cues.
It suggests that an effective solution should enhance head token learning in VFMs while preserving their scene token learning capabilities.

\begin{figure}[t]
    \centering
    \includegraphics[width=\linewidth]{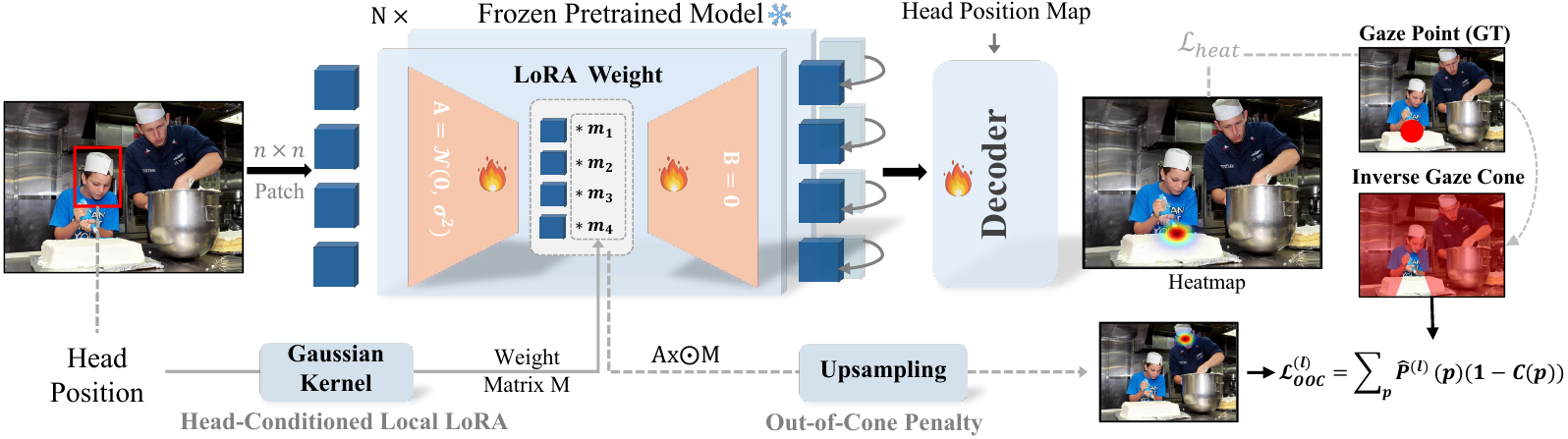}
    \caption{
    Overview of our proposed framework.
    Given an input image and the target person's head box, we construct a head-position prior to modulate the low-rank residual branch inside a frozen pre-trained vision backbone.
    The proposed head-conditioned local LoRA selectively adapts head-relevant features while preserving scene-level representations.
    The adapted features are decoded into the final gaze heatmap, with an additional lightweight in/out branch omitted for clarity.
    During training, auxiliary gaze-evidence maps from intermediate adapted layers are regularized by the proposed out-of-cone penalty, which suppresses prediction mass outside the geometrically plausible gaze cone.
    }
    \label{fig:method}
\end{figure}

\textbf{HCLoRA:} In this work, we propose Head-Conditioned Local LoRA (HCLoRA), a method that adapts the model in a spatially localized manner by focusing the trainable update on the target head region while preserving scene-level representations.
To this end, we build on LoRA, which keeps the pre-trained projection frozen and introduces a lightweight trainable residual branch.
Different from standard LoRA, HCLoRA conditions this residual branch on the target head position, so that the adaptation is emphasized on head tokens and has limited influence on the global scene representation.

In detail, let $\mathbf{I}$ denote the input image and let $b=(x_{\min},y_{\min},x_{\max},y_{\max})$ denote the bounding box of the target head.
Given an input token sequence $\mathbf{X} \in \mathbb{R}^{N \times d}$ and a frozen pre-trained projection $W(\cdot)$, the adapted projection is formulated as
\begin{equation}
\mathbf{Y} = W(\mathbf{X}) + \Delta(\mathbf{X}, b),
\end{equation}
where the head-conditioned residual update is defined as
\begin{equation}
\Delta(\mathbf{X}, b)
=
\bigl(A(\mathbf{X}) \odot \mathbf{M}(b)\bigr)\mathbf{B}.
\end{equation}
Here, $A(\mathbf{X})=\mathbf{X}\mathbf{A} \in \mathbb{R}^{N \times r}$ denotes the low-rank feature produced by the LoRA down-projection, where $\mathbf{A} \in \mathbb{R}^{d \times r}$ and $\mathbf{B} \in \mathbb{R}^{r \times d}$ are trainable low-rank matrices with $r \ll d$.
The matrix $\mathbf{M}(b) \in \mathbb{R}^{N \times r}$ is generated from the target head position and controls where the residual update is applied.

Specifically, we first construct a soft spatial prior from the target head bounding box.
Let $(c_x,c_y)$ denote the box center and $(s_x,s_y)$ denote its scale on the feature map.
For each spatial location $(u,v)$, the head-guided map is computed as
\begin{equation}
H(u,v)=\exp\left(
-\frac{(u-c_x)^2}{2s_x^2+\epsilon}
-\frac{(v-c_y)^2}{2s_y^2+\epsilon}
\right).
\end{equation}
The head-guided map is then passed through a lightweight gating network to produce $\mathbf{M}(b)$.
This modulation matrix is applied to the low-rank feature $A(\mathbf{X})$ before the LoRA up-projection $\mathbf{B}$.
In this way, HCLoRA localizes the trainable adaptation to head-related regions.

\subsection{Out-of-Cone Penalty}
HCLoRA enhances head token extraction, but lacks explicit constraints on what information the head tokens should encode. Encoding gaze cues into head tokens can improve gaze reasoning ability.

\textbf{Misalignment between Head and Scene Tokens.}
A straightforward solution is to supervise head token learning by regressing gaze direction. However, directly regressing gaze direction does not consistently improve gaze-following performance (see Table~\ref{tab:ablation_gaze_sup} for details). We argue that this is due to the misalignment between head tokens and scene tokens. Gaze following methods typically estimate a probability heatmap to represent the gaze target, \textit{i.e.}, each scene token encodes the likelihood of being the target location. In contrast, head tokens are supervised using gaze direction, requiring the decoder to implicitly learn a mapping from a direction vector to a spatial probability distribution. Such a transformation is non-trivial and difficult to learn.

\noindent
\begin{minipage}{\linewidth}
\begin{minipage}[b]{0.28\linewidth}
\textbf{Rigid Representation.} An alternative is to use a gaze cone for supervision by generating a heatmap from head tokens, where conventional methods convert gaze direction into a cone-shaped attention prior.
However, it imposes overly rigid cone-shaped supervision on representations of head tokens.
\par\vspace{0.1pt}
\end{minipage}\hfill
\begin{minipage}[b]{0.69\linewidth}
\centering
\includegraphics[width=\linewidth]{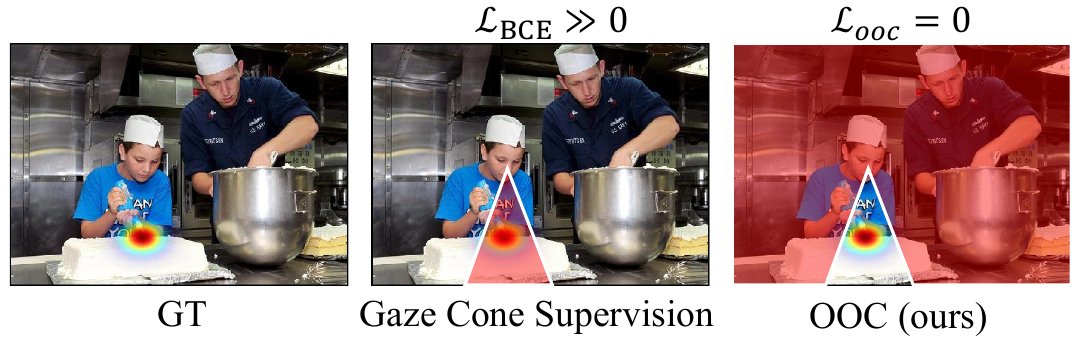}
\makeatletter\def\@captype{figure}\makeatother
\caption{\small Comparison of gaze cone supervision and the out-of-cone penalty. Gaze cone supervision can produce a large loss even when the prediction captures the ground-truth target.}
\label{fig:OOC}
\end{minipage}
\end{minipage}
\medskip

\textbf{OOC Penalty:} In this work, we propose an out-of-cone (OOC) penalty. The OOC penalty has two key strengths:
(1) it leverages head tokens to predict a heatmap aligned with the gaze-following output; and
(2) instead of enforcing rigid gaze-cone supervision, it penalizes probabilities located outside the gaze cone.

This formulation injects gaze cues into head features by discouraging predictions outside the gaze cone rather than enforcing an explicit target distribution. Figure~\ref{fig:OOC} compares gaze-cone supervision with our OOC penalty. Gaze-cone supervision can produce a large loss even when the head token successfully captures the GT, whereas our OOC penalty yields zero penalty in such cases.

Specifically, we attach lightweight auxiliary heads to the last $L$ adapted layers, where $L$ denotes the number of adapted layers with auxiliary supervision.
For each selected layer $l$, given hidden low-rank features $\mathbf{Z}^{(l)} \in \mathbb{R}^{N_p \times r}$, we obtain a normalized auxiliary gaze-evidence map:
\begin{equation}
\hat{\mathbf{P}}^{(l)}
=
\mathrm{Softmax}_{\mathrm{spatial}}
\left(
f_{\mathrm{aux}}^{(l)}(\mathbf{Z}^{(l)})
\right),
\end{equation}
where $f_{\mathrm{aux}}^{(l)}$ maps the low-rank token features to a spatial gaze-evidence map aligned with the supervision resolution.

To penalize auxiliary evidence outside the gaze-consistent region, we construct a soft cone mask centered on the ground-truth gaze direction.
Let $\mathbf{h}$ and $\mathbf{g}$ denote the head center and the ground-truth gaze target, respectively, and let $\mathbf{d}=(\mathbf{g}-\mathbf{h})/(\|\mathbf{g}-\mathbf{h}\|_2+\epsilon)$ denote the normalized gaze direction.
For each spatial location $(u,v)$ with coordinate $\mathbf{p}=\mathbf{x}_{u,v}$, we compute its signed projection on the gaze ray and perpendicular distance to the ray:
\begin{equation}
t(\mathbf{p})=(\mathbf{p}-\mathbf{h})^\top\mathbf{d},\quad
d_\perp(\mathbf{p})=
\left\|(\mathbf{p}-\mathbf{h})-t(\mathbf{p})\mathbf{d}\right\|_2 .
\end{equation}
Given a cone angle $\theta$, the soft cone mask is defined as
\begin{equation}
C(\mathbf{p})
=
\sigma\left(
\alpha\left(
\max(t(\mathbf{p}),0)\tan\frac{\theta}{2}
-
d_\perp(\mathbf{p})
\right)
\right)
\cdot
\sigma\left(\alpha t(\mathbf{p})\right),
\end{equation}
where $\alpha$ controls the boundary sharpness. The first term defines the cone boundary, while the second suppresses locations opposite to the gaze direction.

Finally, we penalize the proportion of auxiliary evidence outside the gaze cone. Since $\sum_{\mathbf{p}}\hat{\mathbf{P}}^{(l)}(\mathbf{p})=1$ by Eq.~(4), the penalty simplifies to:
\begin{equation}
\mathcal{L}_{\mathrm{OOC}}^{(l)}
=
\sum_{\mathbf{p}} \hat{\mathbf{P}}^{(l)}(\mathbf{p}) \bigl(1-C(\mathbf{p})\bigr).
\end{equation}

\subsection{Implementation details}
Our method is built upon the GazeLLE framework~\citep{ryan2025gaze}.
We adopt a DINOv2 backbone with a lightweight gaze decoder, whose output is a gaze heatmap.
The final prediction is supervised using a binary cross-entropy (BCE) loss $\mathcal{L}_{\mathrm{heat}}$.
When required, we additionally use a binary head for in/out prediction, supervised by the BCE loss $\mathcal{L}_{\mathrm{in/out}}$.
The overall training objective is:
\begin{equation}
\mathcal{L}
=
\mathcal{L}_{\mathrm{heat}}
+
\beta \mathcal{L}_{\mathrm{in/out}}
+
\lambda \frac{1}{L}\sum_{l=1}^{L}\mathcal{L}_{\mathrm{OOC}}^{(l)},
\end{equation}
where $\beta$ and $\lambda$ are loss weights.

\begin{table*}[t]
\centering
\setlength{\abovecaptionskip}{0pt}
\setlength{\belowcaptionskip}{7pt}
\caption{
Comparison with state-of-the-art gaze-following methods on GazeFollow.
Our method shows best performance, with particularly large improvements on the \textit{inconsistent} subset.
}
\label{tab:main_gazefollow}
\footnotesize
\setlength{\tabcolsep}{2.1pt}
\renewcommand{\arraystretch}{1.20}
\resizebox{\textwidth}{!}{%
\begin{tabular}{@{}lcccccccccc@{}}
\toprule
\multirow{2}{*}{\textbf{Method}}
& \multicolumn{5}{c}{\textbf{Consistent}}
& \multicolumn{5}{c}{\textbf{Inconsistent}} \\
\cmidrule(lr){2-6} \cmidrule(lr){7-11}
& \textbf{AUC}\upb
& \textbf{Avg. L2}\downb
& \textbf{Min. L2}\downb
& \textbf{Avg. Ang.}\downb
& \textbf{Min. Ang.}\downb
& \textbf{AUC}\upb
& \textbf{Avg. L2}\downb
& \textbf{Min. L2}\downb
& \textbf{Avg. Ang.}\downb
& \textbf{Min. Ang.}\downb \\
\midrule

\citet{chong2020detecting}
& 0.9402 & 0.1194 & 0.0701 & 15.68 & 7.41
& 0.8892 & 0.1731 & 0.0914 & 23.18 & 9.25 \\

\citet{miao2023patch}
& 0.9503 & 0.1044 & 0.0559 & 13.73 & 5.63
& 0.9026 & 0.1601 & 0.0795 & 21.82 & 7.83 \\

Sharingan~\citep{tafasca2024sharingan}
& 0.9586 & 0.0958 & 0.0496 & 10.83 & 4.96
& 0.9148 & 0.1536 & 0.0745 & \underline{17.39} & 7.37 \\

GazeLLE~\citep{ryan2025gaze}
& 0.9689 & 0.0834 & 0.0373 & 11.04 & 3.37
& 0.9270 & 0.1519 & 0.0664 & 21.09 & 7.22 \\

\midrule

GazeLLE + FT
& 0.9613 & 0.0835 & 0.0403 & 11.46 & 3.86
& 0.9183 & 0.1434 & 0.0655 & 20.67 & 7.02 \\

GazeLLE + VPT
& \underline{0.9700} & 0.0825 & 0.0360 & 11.06 & 3.33
& 0.9315 & 0.1527 & 0.0666 & 21.10 & 7.22 \\

GazeLLE + LoRA
& \underline{0.9700} & 0.0789 & 0.0352 & 10.76 & 3.22
& 0.9308 & 0.1487 & 0.0667 & 20.55 & 6.86 \\

\midrule
\rowcolor{gray!15}
\textbf{Ours (ViT-B)}
& 0.9696 & \underline{0.0777} & \underline{0.0348} & \underline{10.50} & \underline{3.19}
& \underline{0.9340} & \underline{0.1353} & \underline{0.0547} & 18.00 & \underline{4.85} \\

\rowcolor{gray!15}
\textbf{Ours (ViT-L)}
& \textbf{0.9719} & \textbf{0.0709} & \textbf{0.0301} & \textbf{9.66} & \textbf{2.58}
& \textbf{0.9371} & \textbf{0.1271} & \textbf{0.0498} & \textbf{17.36} & \textbf{4.44} \\
\bottomrule
\end{tabular}%
}
\end{table*}

\section{Experiments}

\subsection{Experimental Setup}

We evaluate our method on GazeFollow~\citep{recasens2015they} and VideoAttentionTarget (VAT)~\citep{chong2020detecting}. To clearly demonstrate the performance in detail, we use the heuristic strategy described in Sec.~\ref{sec:semantic_partition} to split the evaluation set into \emph{consistent} and \emph{inconsistent} subsets for both datasets. Note that this partition does not modify the evaluation set itself, but only reports results under different subsets.

We use AUC and L2 distance to evaluate gaze-target localization~\citep{recasens2015they,chong2020detecting,ryan2025gaze}.
For VAT, we additionally report AP$_{\mathrm{in/out}}$ for in/out prediction.
In this paper, we introduce angular error as an additional metric for measuring gaze reasoning.
Specifically, given the head center $\mathbf{h}$, the predicted gaze point $\hat{\mathbf{g}}$, and a ground-truth gaze point $\mathbf{g}$, the angular error is computed as
\begin{equation}
\mathrm{Ang.}
=
\arccos
\left(
\frac{
(\hat{\mathbf{g}}-\mathbf{h})^\top(\mathbf{g}-\mathbf{h})
}{
\|\hat{\mathbf{g}}-\mathbf{h}\|_2\|\mathbf{g}-\mathbf{h}\|_2
}
\right).
\end{equation}
This metric measures whether the prediction follows the correct gaze direction from the queried head.

\subsection{Comparison with State-of-the-Art Gaze-Following Methods}

Tables~\ref{tab:main_gazefollow} and~\ref{tab:main_vat} compare our method with state-of-the-art gaze-following methods on GazeFollow and VAT.
On GazeFollow, our method consistently improves over GazeLLE under the same ViT-B backbone, especially on the \emph{inconsistent} subset where semantic saliency is unreliable.
Compared with GazeLLE, our method reduces Avg.~L2 by 10.9\%, Min.~L2 by 17.6\%, Avg.~Ang. by 14.7\%, and Min.~Ang. by 32.8\% on this subset.
It also improves localization accuracy on the \emph{consistent} subset, indicating that stronger gaze reasoning is achieved without sacrificing scene understanding.

The same trend holds on VideoAttentionTarget.
On the \emph{inconsistent} subset, our method reduces L2 error from 0.1670 to 0.1513 and angular error from 24.21 to 18.34.
Compared with full fine-tuning, VPT, and standard LoRA, our method provides more stable gains across datasets and subsets, showing the effectiveness of head-conditioned local adaptation.
Using a stronger ViT-L backbone further improves most localization metrics.
For completeness, an overall comparison on the full test sets with trainable parameter counts is provided in Appendix~\ref{apsota_comparison} and Table~\ref{tab:sota_all}.

\begin{table*}[t]
\centering
\setlength{\abovecaptionskip}{0pt}
\setlength{\belowcaptionskip}{7pt}
\caption{
Comparison with state-of-the-art gaze-following methods on VideoAttentionTarget.
}
\label{tab:main_vat}
\scriptsize
\setlength{\tabcolsep}{2pt}
\renewcommand{\arraystretch}{1.20}
\resizebox{\textwidth}{!}{%
\begin{tabular}{@{}lcccccccccc@{}}
\toprule
\multirow{2}{*}{\textbf{Method}}
& \multicolumn{3}{c}{\textbf{All}}
& \multicolumn{3}{c}{\textbf{Consistent}}
& \multicolumn{3}{c}{\textbf{Inconsistent}}
& \multirow{2}{*}{\textbf{AP}$_{\mathbf{in/out}}$\upb} \\
\cmidrule(lr){2-4} \cmidrule(lr){5-7} \cmidrule(lr){8-10}
& \textbf{AUC}\upb
& \textbf{L2}\downb
& \textbf{Ang.}\downb
& \textbf{AUC}\upb
& \textbf{L2}\downb
& \textbf{Ang.}\downb
& \textbf{AUC}\upb
& \textbf{L2}\downb
& \textbf{Ang.}\downb
& \\
\midrule

\citet{chong2020detecting}
& 0.8628 & 0.1339 & 16.14
& 0.8712 & 0.1237 & 15.56
& 0.8551 & 0.1517 & 17.23
& 0.8510 \\

\citet{miao2023patch}
& 0.9163 & 0.1099 & 12.35
& 0.9295 & 0.0796 & 9.77
& 0.8954 & 0.1580 & \underline{16.45}
& \underline{0.9061} \\

Sharingan~\citep{tafasca2024sharingan}
& 0.9162 & 0.1038 & \underline{12.09}
& 0.9320 & 0.0741 & 9.41
& 0.8912 & \underline{0.1509} & \textbf{16.34}
& 0.8921 \\

GazeLLE~\citep{ryan2025gaze}
& 0.9347 & 0.1071 & 15.02
& 0.9528 & 0.0693 & 9.21
& 0.9059 & 0.1670 & 24.21
& 0.8979 \\

\midrule

GazeLLE + FT
& 0.9347 & 0.1107 & 15.32
& \underline{0.9538} & 0.0746 & 9.60
& 0.9044 & 0.1680 & 24.38
& 0.8858 \\

GazeLLE + VPT
& 0.9309 & 0.1044 & 13.66
& 0.9455 & 0.0709 & 9.31
& \textbf{0.9078} & 0.1576 & 20.56
& 0.8911 \\

GazeLLE + LoRA
& 0.9307 & 0.1077 & 14.24
& 0.9477 & 0.0724 & 9.35
& 0.9037 & 0.1637 & 22.00
& 0.8775 \\

\midrule

\rowcolor{gray!15}
\textbf{Ours (ViT-B)}
& \underline{0.9352} & \underline{0.1006} & 12.26
& 0.9534 & \underline{0.0686} & \underline{8.42}
& 0.9064 & 0.1513 & 18.34
& 0.8988 \\

\rowcolor{gray!15}
\textbf{Ours (ViT-L)}
& \textbf{0.9387} & \textbf{0.0951} & \textbf{11.75}
& \textbf{0.9590} & \textbf{0.0630} & \textbf{7.29}
& \underline{0.9066} & \textbf{0.1459} & 18.83
& \textbf{0.9068} \\

\bottomrule
\end{tabular}%
}
\end{table*}

\begin{table*}[t]
\centering
\setlength{\abovecaptionskip}{9pt}
\setlength{\belowcaptionskip}{7pt}
\caption{
Ablation study on GazeFollow. Both HCLoRA and the OOC penalty improve performance, and reduced angular error indicates enhanced gaze reasoning.
}
\label{tab:ablation_components}
\footnotesize
\setlength{\tabcolsep}{2.8pt}
\renewcommand{\arraystretch}{1.20}
\resizebox{\textwidth}{!}{%
\begin{tabular}{@{}cc ccccc ccccc@{}}
\toprule
\multicolumn{2}{c}{\textbf{Components}}
& \multicolumn{5}{c}{\textbf{Consistent}}
& \multicolumn{5}{c}{\textbf{Inconsistent}} \\
\cmidrule(lr){1-2} \cmidrule(lr){3-7} \cmidrule(lr){8-12}
\textbf{HCLoRA}
& \textbf{OOC penalty}
& \textbf{AUC}\upb
& \textbf{A-L2}\downb
& \textbf{M-L2}\downb
& \textbf{A-Ang}\downb
& \textbf{M-Ang}\downb
& \textbf{AUC}\upb
& \textbf{A-L2}\downb
& \textbf{M-L2}\downb
& \textbf{A-Ang}\downb
& \textbf{M-Ang}\downb \\
\midrule

$\times$ & $\times$
& 0.9689 & 0.0834 & 0.0373 & 11.04 & 3.37
& 0.9270 & 0.1519 & 0.0664 & 21.09 & 7.22 \\

\checkmark & $\times$
& 0.9677 & 0.0786 & 0.0366 & 10.66 & 3.37
& 0.9303 & 0.1383 & 0.0567 & 19.07 & 5.72 \\

\rowcolor{gray!15}
\checkmark & \checkmark
& \textbf{0.9696} & \textbf{0.0777} & \textbf{0.0348} & \textbf{10.50} & \textbf{3.19}
& \textbf{0.9340} & \textbf{0.1353} & \textbf{0.0547} & \textbf{18.00} & \textbf{4.85} \\

\bottomrule
\end{tabular}%
}
\end{table*}

\subsection{Ablation Study}

We conduct component ablation on GazeFollow to examine how the proposed designs contribute to gaze reasoning.
As shown in Table~\ref{tab:ablation_components}, adding HCLoRA leads to substantial gains on the \emph{inconsistent} subset, reducing Avg.~L2 from 0.1519 to 0.1383 and Min.~Ang. from 7.22 to 5.72.
This confirms that localizing the adaptation around head-relevant regions is crucial for improving gaze reasoning while preserving the original VFM representations.
With the out-of-cone penalty, the full model further improves all metrics on the \emph{inconsistent} subset, achieving 0.1353 Avg.~L2 and 4.85 Min.~Ang., while also maintaining strong performance on the \emph{consistent} subset.

Additional ablations on hyperparameter sensitivity and architectural variants are provided in Appendix Tables~\ref{tab:hyper_sensitivity}, \ref{tab:gazelle_design_ablation}, and~\ref{tab:ours_head_scene_branch}.
These results further support that our method robustly improves gaze reasoning through localized adaptation.

\subsection{Comparison with Alternative Gaze-Aware Supervision Strategies}
\begin{table*}[t]
\centering
\setlength{\abovecaptionskip}{0pt}
\setlength{\belowcaptionskip}{7pt}
\caption{
Comparison with other gaze-aware supervision strategies on GazeFollow.
All variants use HCLoRA and differ only in the auxiliary supervision strategy.
}
\label{tab:ablation_gaze_sup}
\footnotesize
\setlength{\tabcolsep}{2.2pt}
\renewcommand{\arraystretch}{1.20}
\resizebox{\textwidth}{!}{%
\begin{tabular}{@{}l ccccc ccccc@{}}
\toprule
\multirow{2}{*}{\textbf{Method}}
& \multicolumn{5}{c}{\textbf{Consistent}}
& \multicolumn{5}{c}{\textbf{Inconsistent}} \\
\cmidrule(lr){2-6} \cmidrule(lr){7-11}
& \textbf{AUC}\upb
& \textbf{A-L2}\downb
& \textbf{M-L2}\downb
& \textbf{A-Ang}\downb
& \textbf{M-Ang}\downb
& \textbf{AUC}\upb
& \textbf{A-L2}\downb
& \textbf{M-L2}\downb
& \textbf{A-Ang}\downb
& \textbf{M-Ang}\downb \\
\midrule

w/o supervision
& 0.9677 & 0.0786 & 0.0366 & 10.66 & 3.37
& 0.9303 & 0.1383 & 0.0567 & 19.07 & 5.72 \\

\midrule

Gaze-vector
& 0.9682 & 0.0790 & 0.0366 & 10.34 & 3.11
& 0.9318 & 0.1378 & 0.0578 & 18.80 & 5.42 \\

Heatmap
& 0.9681 & 0.0782 & 0.0359 & 10.51 & 3.13
& 0.9293 & 0.1385 & 0.0586 & 18.59 & 5.42 \\

Gaze-cone
& 0.9678 & 0.0777 & \textbf{0.0347} & \textbf{10.25} & \textbf{3.00}
& 0.9304 & 0.1385 & 0.0578 & 18.59 & 5.26 \\

\midrule

\rowcolor{gray!15}
Out-of-cone penalty
& \textbf{0.9696} & \textbf{0.0777} & 0.0348 & 10.50 & 3.19
& \textbf{0.9340} & \textbf{0.1353} & \textbf{0.0547} & \textbf{18.00} & \textbf{4.85} \\

\bottomrule
\end{tabular}%
}
\end{table*}

\begin{figure*}[!b]
    \centering
    \includegraphics[width=\linewidth]{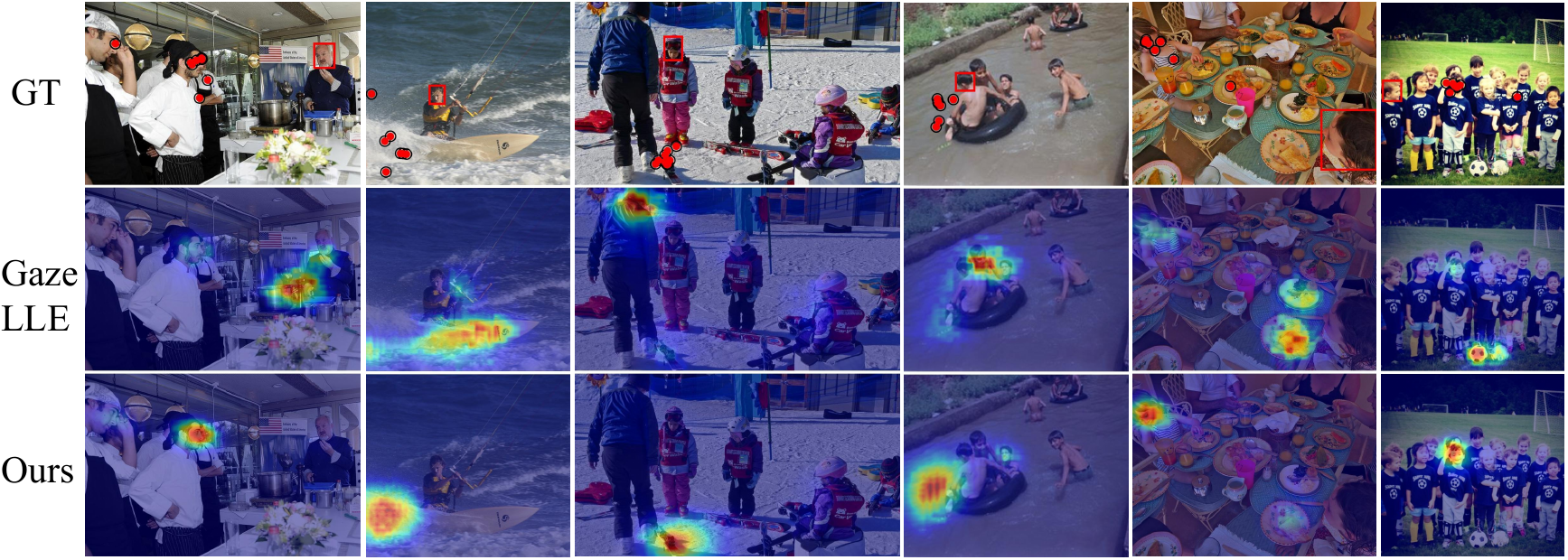}
    \caption{
    We visualize the predictions of GazeLLE and our method for comparison.
    GazeLLE often assigns high responses to semantically salient but incorrect regions, while our method follows the queried person's gaze and localizes the true target.
    This shows improved gaze reasoning beyond semantic saliency.
    Zoom in for facial and gaze direction details.
    }
    \label{fig:qualitative_comparison}
\end{figure*}

We study different forms of gaze-aware supervision in Table~\ref{tab:ablation_gaze_sup}.
Direct gaze-vector regression, heatmap prediction, and rigid gaze-cone supervision bring limited gains.
For example, although gaze-cone supervision improves several metrics on the \emph{consistent} subset, it does not reduce the Avg.~L2 error on the \emph{inconsistent} subset compared with no auxiliary supervision.
In contrast, the proposed out-of-cone penalty achieves the best performance on the \emph{inconsistent} subset, improving AUC from 0.9303 to 0.9340, reducing Avg.~L2 from 0.1383 to 0.1353, and reducing Min.~Ang. from 5.72 to 4.85.
It also remains competitive on the \emph{consistent} subset, achieving the best AUC and tied best Avg.~L2.
These results indicate that softly suppressing gaze-inconsistent evidence is more effective than forcing intermediate features to match a fixed direction vector or rigid cone-shaped distribution.

\medskip
\newsavebox{\gazeanalysisbox}
\begin{lrbox}{\gazeanalysisbox}
\begin{minipage}{\linewidth}
\subsection{Gaze Reasoning Ability Analysis}
\begin{minipage}[t]{0.43\linewidth}
\vspace{0pt}
Our goal is to improve the gaze reasoning ability. Compared with the L2 error, angular error better reflects gaze reasoning capability. Therefore, we conduct experiments to count the number of samples with large angular errors. Intuitively, methods with fewer large-error samples exhibit stronger \mbox{gaze reasoning ability}.

\looseness=1
The results are shown in Fig.~\ref{fig:large_error_tail}. “$>k^\circ$” denotes the number of samples whose minimum angular error exceeds $k^\circ$. Our method is built upon GazeLLE. The results show that both HCLoRA and OOC improve gaze reasoning ability.
Under the $30^\circ$ threshold, the number of large-error samples decreases from 96 to 70 with HCLoRA, and further to 52 with OOC. These results indicate that our method not only improves overall performance, but also enhances gaze reasoning ability, consistent with our \mbox{original motivation}.
\end{minipage}\hfill
\begin{minipage}[t]{0.54\linewidth}
\vspace{0pt}
\centering
\addtolength{\abovecaptionskip}{6.1pt}%
\includegraphics[width=\linewidth]{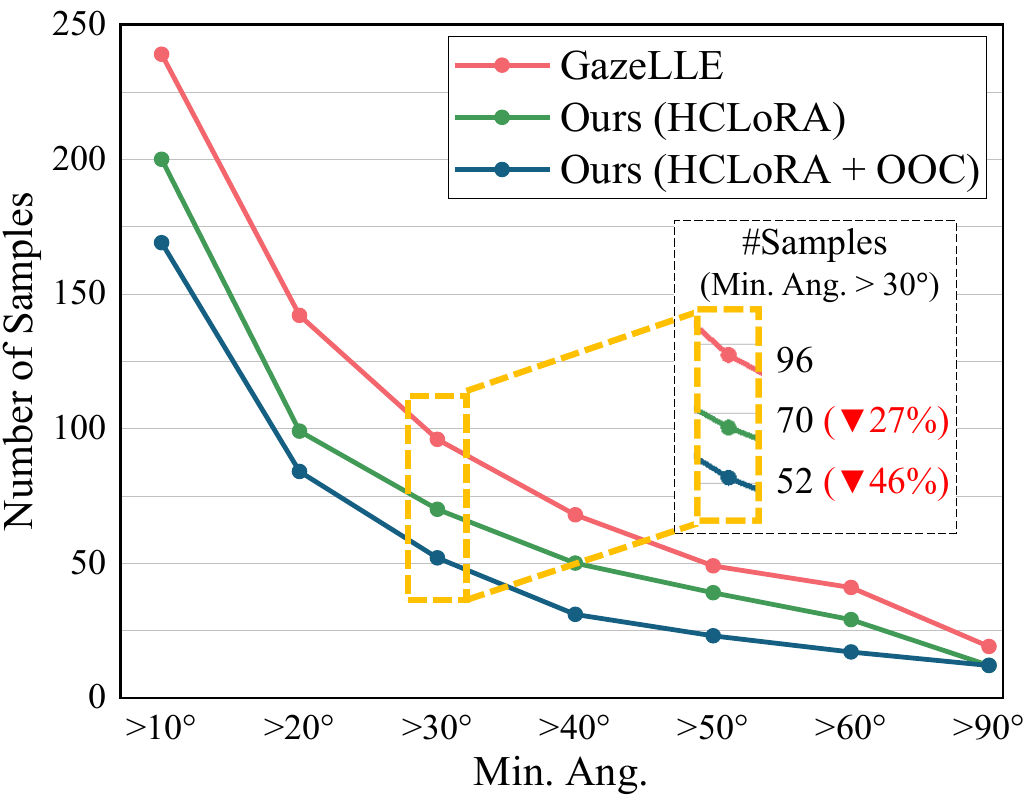}
\makeatletter\def\@captype{figure}\makeatother
\caption{We count the number of samples with large angular errors. “$>k^\circ$” denotes the number of samples whose minimum angular error exceeds $k^\circ$. The results show that our method effectively enhances gaze reasoning ability, consistent with our motivation.
}
\label{fig:large_error_tail}
\end{minipage}
\end{minipage}
\end{lrbox}

\begin{figure}[!b]
    \centering
    \includegraphics[width=0.95\linewidth]{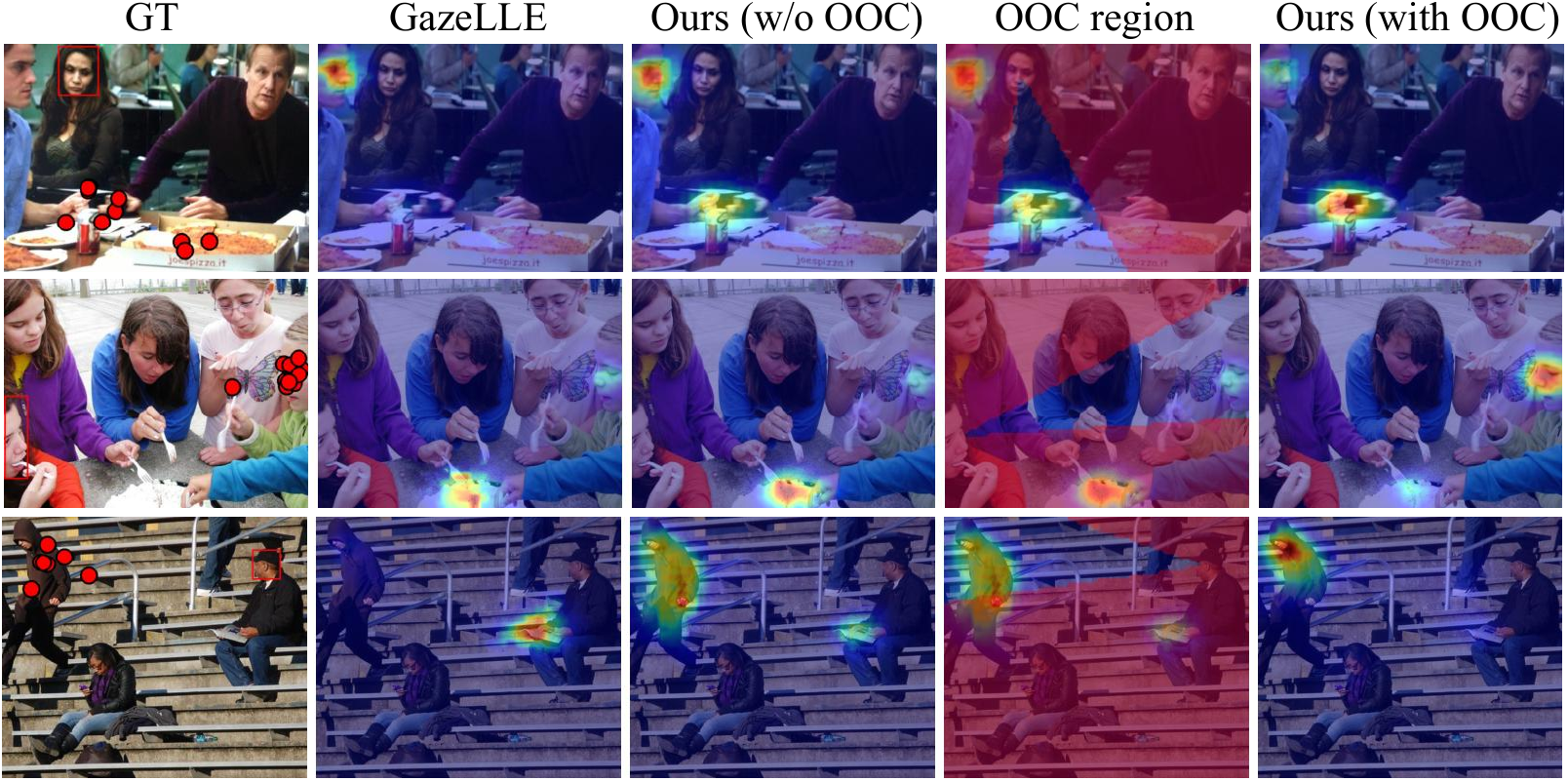}
    \caption{
    We visualize the impact of the OOC penalty on gaze prediction. The OOC penalty improves performance by suppressing predictions outside the gaze cone.}
    \label{fig:qualitative_ooc}
\end{figure}

\noindent\usebox{\gazeanalysisbox}

\subsection{Visualization Results}

We further provide qualitative visualizations to demonstrate the advantage of our method.
As shown in Fig.~\ref{fig:qualitative_comparison}, GazeLLE often produces high responses around semantically salient objects or interaction centers, even when these regions are inconsistent with the queried person's gaze direction.
In contrast, our method more accurately follows the target person's head and gaze cues, shifting the response toward the true gaze target.
This shows that our method reduces reliance on semantic saliency and improves localization of the correct gaze direction.

Fig.~\ref{fig:qualitative_ooc} visualizes the effect of the out-of-cone penalty.
We overlay the OOC region on the prediction of HCLoRA \textit{w/o} OOC.
The OOC penalty suppresses these gaze-inconsistent responses and concentrates more clearly on the gaze-consistent target.
It guides the model toward more accurate target localization by discouraging evidence outside the gaze-consistent region.

\begin{samepage}
\section{Conclusion}

We study the limitations of VFM-based gaze following and show that strong benchmark performance can still be driven by semantic shortcuts rather than reliable gaze reasoning.
To address this, we propose HCLoRA and OOC penalty to enhance gaze reasoning.
Experiments on GazeFollow and VideoAttentionTarget demonstrate that our method achieves state-of-the-art performance while significantly improving gaze reasoning ability, consistent with our primary motivation.

\end{samepage}

\renewcommand{\acksection}{\section*{Acknowledgments}}
\begin{ack}
This work was supported by the National Natural Science Foundation of China (62672028, 62271042, 62376021), the Beijing Natural Science Foundation (L259055), and the Institute of Information \& Communications Technology Planning \& Evaluation (IITP) grant funded by the Korean government (MSIT) (No. RS-2022-II220608). The research was conducted during Shijing Wang's visit to the University of Birmingham.
\end{ack}

\bibliographystyle{plainnat}
\bibliography{main}

\begin{thebibliography}{30}
\providecommand{\natexlab}[1]{#1}
\providecommand{\url}[1]{\texttt{#1}}
\expandafter\ifx\csname urlstyle\endcsname\relax
  \providecommand{\doi}[1]{doi: #1}\else
  \providecommand{\doi}{doi: \begingroup \urlstyle{rm}\Url}\fi

\bibitem[Admoni and Scassellati(2017)]{admoni2017social}
Henny Admoni and Brian Scassellati.
\newblock Social eye gaze in human-robot interaction: a review.
\newblock \emph{Journal of Human-Robot Interaction}, 6\penalty0 (1):\penalty0 25--63, 2017.

\bibitem[Bachmann et~al.(2022)Bachmann, Mizrahi, Atanov, and Zamir]{bachmann2022multimae}
Roman Bachmann, David Mizrahi, Andrei Atanov, and Amir Zamir.
\newblock {MultiMAE}: Multi-modal multi-task masked autoencoders.
\newblock In \emph{European Conference on Computer Vision}, pages 348--367. Springer Nature Switzerland, 2022.

\bibitem[Bai et~al.(2025)Bai, Cai, Chen, Chen, Chen, Cheng, Deng, Ding, Gao, Ge, et~al.]{bai2025qwen3}
Shuai Bai, Yuxuan Cai, Ruizhe Chen, Keqin Chen, Xionghui Chen, Zesen Cheng, Lianghao Deng, Wei Ding, Chang Gao, Chunjiang Ge, et~al.
\newblock {Qwen3-VL} technical report.
\newblock \emph{arXiv preprint arXiv:2511.21631}, 2025.

\bibitem[Bao et~al.(2022)Bao, Liu, and Yu]{bao2022escnet}
Jun Bao, Buyu Liu, and Jun Yu.
\newblock {ESCNet}: Gaze target detection with the understanding of {3D} scenes.
\newblock In \emph{Proceedings of the IEEE/CVF Conference on Computer Vision and Pattern Recognition}, pages 14106--14115, 2022.

\bibitem[Capozzi et~al.(2019)Capozzi, Beyan, Pierro, Koul, Murino, Livi, Bayliss, Ristic, and Becchio]{capozzi2019tracking}
Francesca Capozzi, Cigdem Beyan, Antonio Pierro, Atesh Koul, Vittorio Murino, Stefano Livi, Andrew~P. Bayliss, Jelena Ristic, and Cristina Becchio.
\newblock Tracking the leader: Gaze behavior in group interactions.
\newblock \emph{{iScience}}, 16:\penalty0 242--249, 2019.

\bibitem[Chac{\'o}n~Quesada et~al.(2025)Chac{\'o}n~Quesada, Est{\'e}vez~Casado, and Demiris]{quesada2025integrated}
Rodrigo Chac{\'o}n~Quesada, Fernando Est{\'e}vez~Casado, and Yiannis Demiris.
\newblock An integrated {3D} eye-gaze tracking framework for assessing trust in human--robot interaction.
\newblock \emph{ACM Transactions on Human-Robot Interaction}, 14\penalty0 (3):\penalty0 57:1--57:28, 2025.

\bibitem[Chen et~al.(2022)Chen, Xu, Zhu, Liu, Lu, Zheng, and Kong]{chen2021gaze}
Wenhe Chen, Hui Xu, Chao Zhu, Xiaoli Liu, Yinghua Lu, Caixia Zheng, and Jun Kong.
\newblock Gaze estimation via the joint modeling of multiple cues.
\newblock \emph{IEEE Transactions on Circuits and Systems for Video Technology}, 32\penalty0 (3):\penalty0 1390--1402, 2022.

\bibitem[Chong et~al.(2018)Chong, Ruiz, Wang, Zhang, Rozga, and Rehg]{chong2018connecting}
Eunji Chong, Nataniel Ruiz, Yongxin Wang, Yun Zhang, Agata Rozga, and James~M. Rehg.
\newblock Connecting gaze, scene, and attention: Generalized attention estimation via joint modeling of gaze and scene saliency.
\newblock In \emph{Proceedings of the European Conference on Computer Vision (ECCV)}, pages 397--412, 2018.

\bibitem[Chong et~al.(2020)Chong, Wang, Ruiz, and Rehg]{chong2020detecting}
Eunji Chong, Yongxin Wang, Nataniel Ruiz, and James~M. Rehg.
\newblock Detecting attended visual targets in video.
\newblock In \emph{Proceedings of the IEEE/CVF Conference on Computer Vision and Pattern Recognition}, pages 5395--5405, 2020.

\bibitem[Fang et~al.(2021)Fang, Tang, Shen, Shen, Gu, Song, and Zhai]{fang2021dual}
Yi~Fang, Jiapeng Tang, Wang Shen, Wei Shen, Xiao Gu, Li~Song, and Guangtao Zhai.
\newblock Dual attention guided gaze target detection in the wild.
\newblock In \emph{Proceedings of the IEEE/CVF Conference on Computer Vision and Pattern Recognition}, pages 11390--11399, 2021.

\bibitem[Gupta et~al.(2022)Gupta, Tafasca, and Odobez]{gupta2022modular}
Anshul Gupta, Samy Tafasca, and Jean-Marc Odobez.
\newblock A modular multimodal architecture for gaze target prediction: Application to privacy-sensitive settings.
\newblock In \emph{Proceedings of the IEEE/CVF Conference on Computer Vision and Pattern Recognition (CVPR) Workshops}, pages 5037--5046, 2022.

\bibitem[Gupta et~al.(2024)Gupta, Vuillecard, Farkhondeh, and Odobez]{gupta2024exploring}
Anshul Gupta, Pierre Vuillecard, Arya Farkhondeh, and Jean-Marc Odobez.
\newblock Exploring the zero-shot capabilities of vision-language models for improving gaze following.
\newblock In \emph{Proceedings of the IEEE/CVF Conference on Computer Vision and Pattern Recognition (CVPR) Workshops}, pages 615--624, 2024.

\bibitem[Hessels et~al.(2025)Hessels, Iwabuchi, Niehorster, Funawatari, Benjamins, Kawakami, Nystr{\"o}m, Suda, Hooge, Sumiya, et~al.]{hessels2025gaze}
Roy~S. Hessels, Toshiki Iwabuchi, Diederick~C. Niehorster, Ren Funawatari, Jeroen~S. Benjamins, Sayaka Kawakami, Marcus Nystr{\"o}m, Momoka Suda, Ignace T.~C. Hooge, Motofumi Sumiya, et~al.
\newblock Gaze behavior in face-to-face interaction: A cross-cultural investigation between {Japan} and {The Netherlands}.
\newblock \emph{Cognition}, 263:\penalty0 106174, 2025.

\bibitem[Horanyi et~al.(2023)Horanyi, Zheng, Chong, Leonardis, and Chang]{horanyi2023they}
Nora Horanyi, Linfang Zheng, Eunji Chong, Ale{\v{s}} Leonardis, and Hyung~Jin Chang.
\newblock Where are they looking in the {3D} space?
\newblock In \emph{Proceedings of the IEEE/CVF Conference on Computer Vision and Pattern Recognition (CVPR) Workshops}, pages 2678--2687, 2023.

\bibitem[Hu et~al.(2022{\natexlab{a}})Hu, Shen, Wallis, Allen-Zhu, Li, Wang, Wang, and Chen]{hu2021lora}
Edward~J. Hu, Yelong Shen, Phillip Wallis, Zeyuan Allen-Zhu, Yuanzhi Li, Shean Wang, Lu~Wang, and Weizhu Chen.
\newblock {LoRA}: Low-rank adaptation of large language models.
\newblock In \emph{International Conference on Learning Representations (ICLR)}, 2022{\natexlab{a}}.

\bibitem[Hu et~al.(2022{\natexlab{b}})Hu, Zhao, Zhou, Guo, Wu, Yang, and Liu]{hu2022gaze}
Zhengxi Hu, Kunxu Zhao, Bohan Zhou, Hang Guo, Shichao Wu, Yuxue Yang, and Jingtai Liu.
\newblock Gaze target estimation inspired by interactive attention.
\newblock \emph{IEEE Transactions on Circuits and Systems for Video Technology}, 32\penalty0 (12):\penalty0 8524--8536, 2022{\natexlab{b}}.

\bibitem[Jin et~al.(2022)Jin, Yu, Zhu, Lin, Ren, Zhou, and Song]{jin2022depth}
Tianlei Jin, Qizhi Yu, Shiqiang Zhu, Zheyuan Lin, Jie Ren, Yuanhai Zhou, and Wei Song.
\newblock Depth-aware gaze-following via auxiliary networks for robotics.
\newblock \emph{Engineering Applications of Artificial Intelligence}, 113:\penalty0 104924, 2022.

\bibitem[Kim and Lee(2021)]{kim2021assessing}
Nayeon Kim and Hyunsoo Lee.
\newblock Assessing consumer attention and arousal using eye-tracking technology in virtual retail environment.
\newblock \emph{Frontiers in Psychology}, 12:\penalty0 665658, 2021.

\bibitem[Li et~al.(2022)Li, Chen, Zhong, Lam, Han, Ouyang, Li, and Liu]{li2022appearance}
Jing Li, Zejin Chen, Yihao Zhong, Hak-Keung Lam, Junxia Han, Gaoxiang Ouyang, Xiaoli Li, and Honghai Liu.
\newblock Appearance-based gaze estimation for {ASD} diagnosis.
\newblock \emph{IEEE Transactions on Cybernetics}, 52\penalty0 (7):\penalty0 6504--6517, 2022.

\bibitem[Lian et~al.(2018)Lian, Yu, and Gao]{lian2018believe}
Dongze Lian, Zehao Yu, and Shenghua Gao.
\newblock Believe it or not, we know what you are looking at!
\newblock In \emph{Asian Conference on Computer Vision}, pages 35--50, 2018.

\bibitem[Miao et~al.(2023)Miao, Hoai, and Samaras]{miao2023patch}
Qiaomu Miao, Minh Hoai, and Dimitris Samaras.
\newblock Patch-level gaze distribution prediction for gaze following.
\newblock In \emph{Proceedings of the IEEE/CVF Winter Conference on Applications of Computer Vision}, pages 880--889, 2023.

\bibitem[Oquab et~al.(2024)Oquab, Darcet, Moutakanni, Vo, Szafraniec, Khalidov, Fernandez, Haziza, Massa, El-Nouby, Assran, Ballas, Galuba, Howes, Huang, Li, Misra, Rabbat, Sharma, Synnaeve, Xu, J{\'e}gou, Mairal, Labatut, Joulin, and Bojanowski]{oquab2023dinov2}
Maxime Oquab, Timoth{\'e}e Darcet, Th{\'e}o Moutakanni, Huy~V. Vo, Marc Szafraniec, Vasil Khalidov, Pierre Fernandez, Daniel Haziza, Francisco Massa, Alaaeldin El-Nouby, Mahmoud Assran, Nicolas Ballas, Wojciech Galuba, Russell Howes, Po-Yao Huang, Shang-Wen Li, Ishan Misra, Michael Rabbat, Vasu Sharma, Gabriel Synnaeve, Hu~Xu, Herv{\'e} J{\'e}gou, Julien Mairal, Patrick Labatut, Armand Joulin, and Piotr Bojanowski.
\newblock {DINOv2}: Learning robust visual features without supervision.
\newblock \emph{Transactions on Machine Learning Research}, 2024.
\newblock URL \url{https://openreview.net/forum?id=a68SUt6zFt}.

\bibitem[Recasens et~al.(2015)Recasens, Khosla, Vondrick, and Torralba]{recasens2015they}
Adri{\`a} Recasens, Aditya Khosla, Carl Vondrick, and Antonio Torralba.
\newblock Where are they looking?
\newblock In \emph{Advances in Neural Information Processing Systems}, volume~28, pages 199--207, 2015.

\bibitem[Ryan et~al.(2025)Ryan, Bati, Lee, Bolya, Hoffman, and Rehg]{ryan2025gaze}
Fiona Ryan, Ajay Bati, Sangmin Lee, Daniel Bolya, Judy Hoffman, and James~M. Rehg.
\newblock {Gaze-LLE}: Gaze target estimation via large-scale learned encoders.
\newblock In \emph{Proceedings of the IEEE/CVF Conference on Computer Vision and Pattern Recognition (CVPR)}, pages 28874--28884, 2025.

\bibitem[Tafasca et~al.(2023{\natexlab{a}})Tafasca, Gupta, Kojovic, Gelsomini, Maillart, Papandrea, Schaer, and Odobez]{tafasca2023ai4autism}
Samy Tafasca, Anshul Gupta, Nada Kojovic, Mirko Gelsomini, Thomas Maillart, Michela Papandrea, Marie Schaer, and Jean-Marc Odobez.
\newblock The {AI4Autism} project: A multimodal and interdisciplinary approach to autism diagnosis and stratification.
\newblock In \emph{Companion Publication of the 25th International Conference on Multimodal Interaction}, pages 414--425, 2023{\natexlab{a}}.

\bibitem[Tafasca et~al.(2023{\natexlab{b}})Tafasca, Gupta, and Odobez]{tafasca2023childplay}
Samy Tafasca, Anshul Gupta, and Jean-Marc Odobez.
\newblock {ChildPlay}: A new benchmark for understanding children's gaze behaviour.
\newblock In \emph{Proceedings of the IEEE/CVF International Conference on Computer Vision}, pages 20878--20889, 2023{\natexlab{b}}.

\bibitem[Tafasca et~al.(2024)Tafasca, Gupta, and Odobez]{tafasca2024sharingan}
Samy Tafasca, Anshul Gupta, and Jean-Marc Odobez.
\newblock {Sharingan}: A transformer architecture for multi-person gaze following.
\newblock In \emph{Proceedings of the IEEE/CVF Conference on Computer Vision and Pattern Recognition}, pages 2008--2017, 2024.

\bibitem[Tonini et~al.(2023)Tonini, Dall'Asen, Beyan, and Ricci]{tonini2023object}
Francesco Tonini, Nicola Dall'Asen, Cigdem Beyan, and Elisa Ricci.
\newblock Object-aware gaze target detection.
\newblock In \emph{Proceedings of the IEEE/CVF International Conference on Computer Vision}, pages 21803--21812, 2023.

\bibitem[Tu et~al.(2022)Tu, Min, Duan, Guo, Zhai, and Shen]{tu2022end}
Danyang Tu, Xiongkuo Min, Huiyu Duan, Guodong Guo, Guangtao Zhai, and Wei Shen.
\newblock End-to-end human-gaze-target detection with transformers.
\newblock In \emph{2022 IEEE/CVF Conference on Computer Vision and Pattern Recognition (CVPR)}, pages 2192--2200. IEEE, 2022.

\bibitem[Zhang et~al.(2025)Zhang, Yang, Chen, Shi, Wang, Luo, and Huang]{zhang2025mindeye}
Zejia Zhang, Bo~Yang, Xinxing Chen, Weizhuang Shi, Haoyuan Wang, Wei Luo, and Jian Huang.
\newblock {MindEye-OmniAssist}: A gaze-driven {LLM}-enhanced assistive robot system for implicit intention recognition and task execution.
\newblock In \emph{2025 IEEE International Conference on Cyborg and Bionic Systems (CBS)}, pages 1--6. IEEE, 2025.

\end{thebibliography}

\par\vspace{28pt}
\appendix

\section{Additional Details for Observation and Analysis}
\label{app:observation_analysis}

In this appendix, we provide additional details for the observation and analysis in Section~\ref{sec:semantic_partition}.
We first present more failure cases that illustrate the semantic shortcut behavior discussed in Section~\ref{sec:semantic_partition}.
We then describe the prompt used for semantic target annotation and provide qualitative examples of the semantic-consistency-based partition.

\subsection{More Failure Cases of Semantic Shortcut}
\label{app:semantic_shortcut_failures}

\begin{figure*}[t]
    \centering
    \includegraphics[width=\linewidth]{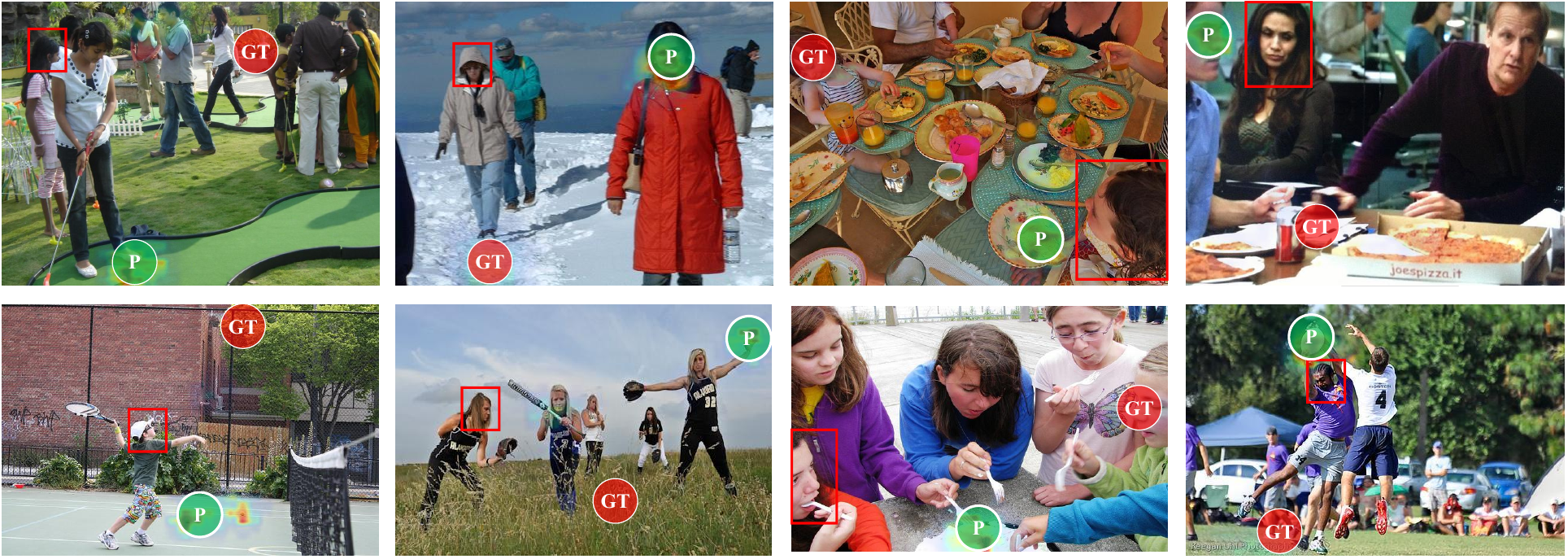}
    \caption{
    Additional failure cases of semantic shortcut behavior in GazeLLE~\citep{ryan2025gaze}.
    The predictions often concentrate on semantically salient regions or interaction centers, while deviating from the true gaze targets.
    These cases further illustrate that existing VFM-based gaze-following methods may rely on semantic saliency rather than robust gaze reasoning.
    }
    \label{fig:semantic_shortcut_failures}
\end{figure*}

Figure~\ref{fig:semantic_shortcut_failures} shows additional failure cases of GazeLLE~\citep{ryan2025gaze}.
These examples further support our observation that the model often predicts semantically salient regions, such as manipulated objects or interaction centers, even when these regions are inconsistent with the target person's gaze direction.
This behavior indicates that the model may rely on semantic plausibility rather than true gaze cues.

\subsection{Prompt for Semantic Target Annotation}
\label{app:semantic_partition_details}

We use Qwen3-VL-32B-Instruct~\citep{bai2025qwen3} as a semantic prober to annotate contextually plausible gaze regions for the target person.
The input image contains a red bounding box marking the target person's head.
The model is asked to return four types of semantic regions: hand object, social target, action target, and semantic center.
The full prompt is shown below.

\begingroup
\tcbset{breakable, lines before break=8}
\begin{promptbox}{Prompt for Semantic Target Annotation}
\small
\textbf{Preamble} \\
\texttt{<Task Description> \newline
The red box marks the HEAD of the target person. \newline
\newline
Your job is to identify semantic targets in the scene that are strongly related to what this person may be attending to. \newline
Return results for the following four slots:}

\vspace{3mm}
\textbf{Semantic Slots} \\
\texttt{<Slots> \newline
1. hand\_object: \newline
\hspace*{1em}ONE non-person object that this person is holding, touching with the hands, or directly operating right now. \newline
\hspace*{1em}Examples: phone, cup, laptop, book, tool, remote controller, steering wheel, toy. \newline
\newline
2. social\_target: \newline
\hspace*{1em}ONE other person who is directly socially interacting with the target person. \newline
\hspace*{1em}Examples: the person they are talking to, hugging, facing in conversation, handing something to, or responding to. \newline
\newline
3. action\_target: \newline
\hspace*{1em}ONE object that is the direct target of this person's current action. \newline
\hspace*{1em}Examples: the ball they just hit or throw, the skateboard they are riding, \newline
\hspace*{1em}the obstacle they are jumping onto, the door they are opening, the screen they are typing on. \newline
\newline
4. semantic\_center: \newline
\hspace*{1em}ONE event-central object/person in the scene that is most semantically important for understanding \newline
\hspace*{1em}what is happening, and is also plausibly something the target person may attend to. \newline
\hspace*{1em}Examples: birthday cake at a birthday party, ball in a sports play, microphone or podium in a speech, \newline
\hspace*{1em}board or screen in a classroom, gift being opened, musical instrument being performed.}

\tcbbreak
\textbf{Instructions} \\
\texttt{<Requirements> \newline
1. Do NOT return the target person themselves. \newline
2. Do NOT choose generic background regions or random furniture unless clearly central. \newline
3. If a slot does not exist, return null for that slot.}

\vspace{3mm}
\textbf{Output Format} \\
\texttt{<JSON Schema> \newline
Return ONLY valid JSON in this exact schema: \newline
\{ \newline
\hspace*{1em}"hand\_object": \{"bbox\_2d": [x1, y1, x2, y2], "label": "..." \} or null, \newline
\hspace*{1em}"social\_target": \{"bbox\_2d": [x1, y1, x2, y2], "label": "..." \} or null, \newline
\hspace*{1em}"action\_target": \{"bbox\_2d": [x1, y1, x2, y2], "label": "..." \} or null, \newline
\hspace*{1em}"semantic\_center": \{"bbox\_2d": [x1, y1, x2, y2], "label": "..." \} or null \newline
\} \newline
\newline
Coordinates must be integers from 0 to 1000 and use [x1, y1, x2, y2].}
\end{promptbox}
\endgroup

\subsection{Qualitative Examples of Semantic-Consistency-Based Partition}
\label{app:semantic_partition_examples}

\begin{figure}[!ht]
    \centering
    \includegraphics[width=\linewidth]{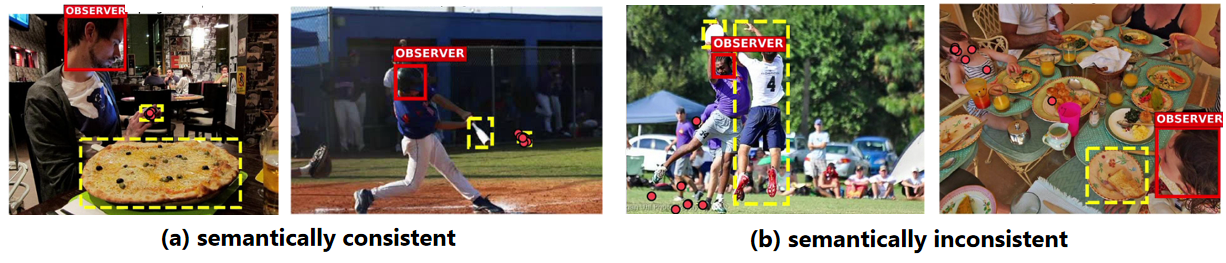}
    \caption{
    Semantic-consistency-based partition examples.
    The red box marks the target person's head, yellow dashed boxes denote semantic regions predicted by the semantic prober, and red circles denote ground-truth gaze annotations.
    \textbf{(a)} Consistent samples, where most gaze annotations fall inside the predicted semantic regions.
    \textbf{(b)} Inconsistent samples, where gaze annotations lie outside the predicted semantic regions despite their contextual saliency.
    }
    \label{fig:semantic_partition_examples}
\end{figure}

Figure~\ref{fig:semantic_partition_examples} provides representative examples of the semantic-consistency-based partition.
For each sample, the semantic prober predicts contextually plausible gaze regions, shown as yellow dashed bounding boxes.
The ground-truth gaze annotations are overlaid as red circles.
A sample is labeled as \emph{consistent} if more than half of the valid ground-truth gaze annotations fall inside at least one predicted semantic region; otherwise, it is labeled as \emph{inconsistent}.

\FloatBarrier

\section{Additional Experimental Results}
\label{app:additional_experimental_results}

\subsection{Additional Experimental Details}
\label{app:experimental_details}

\paragraph{Dataset partition.}
We provide the exact subset statistics used in our semantic-consistency-based evaluation.
GazeFollow contains only in-frame test samples, which are divided into 3,179 \emph{consistent} and 1,603 \emph{inconsistent} samples.
For VideoAttentionTarget (VAT), we apply the semantic-consistency partition to in-frame samples and keep out-of-frame samples as a separate group, obtaining 12,664 \emph{consistent}, 7,991 \emph{inconsistent}, and 11,323 out-of-frame samples.
These subset-level statistics are used for the diagnostic evaluation reported in the main experiments.

\paragraph{Implementation details.}
The input image resolution is set to $448 \times 448$.
For GazeFollow, we train the model for 15 epochs with a batch size of 60.
We use the Adam optimizer with an initial learning rate of $1\times10^{-3}$ and a cosine learning-rate schedule.
For VideoAttentionTarget, we initialize from the GazeFollow-trained model and fine-tune for 8 epochs.
Specifically, we first train only the in/out prediction parameters for 3 epochs with a learning rate of $1\times10^{-2}$.
We then train all non-backbone parameters for another 5 epochs, using a learning rate of $1\times10^{-7}$ for the HCLoRA modules and $1\times10^{-5}$ for the remaining decoder parameters.

\paragraph{HCLoRA-specific settings.}
For HCLoRA, the gaze-aware supervision loss is applied to the last $L=4$ adapted layers with a weight of $\lambda=0.02$.
For the soft forward gaze cone, we use a cone angle of $30^\circ$ and set the sharpness parameter to $\alpha=40$.

\paragraph{Comparison methods.}
We compare HCLoRA with representative gaze-following methods, including \citet{chong2020detecting}, \citet{miao2023patch}, Sharingan~\citep{tafasca2024sharingan}, and GazeLLE~\citep{ryan2025gaze}.
These methods cover both earlier CNN/transformer-based models and recent approaches built on large pre-trained vision backbones.
In particular, Sharingan and GazeLLE are strong recent baselines with highly competitive performance on standard gaze-following benchmarks, making them suitable for evaluating robustness under semantic shortcuts.
Since HCLoRA is built on the GazeLLE framework, we use GazeLLE as the main baseline and further compare with adaptation variants using the same backbone, including full fine-tuning, standard LoRA, and visual prompt tuning (VPT).
For full fine-tuning on GazeFollow, we start from a learning rate of $1\times10^{-5}$, as the default learning rate is too large for updating all backbone parameters and leads to unstable training.

\subsection{Overall Comparison with State-of-the-Art Methods}
\label{apsota_comparison}

Table~\ref{tab:sota_all} summarizes the overall comparison with prior state-of-the-art methods on GazeFollow and VideoAttentionTarget.
Results of prior methods are taken from GazeLLE~\citep{ryan2025gaze}, while our overall results are computed from the corresponding subset-level evaluations.
Our method achieves the best overall localization performance on both benchmarks using only RGB input.
Compared with GazeLLE under the same ViT-B backbone, our method reduces Avg.~L2 from 0.104 to 0.097 on GazeFollow and L2 from 0.107 to 0.101 on VideoAttentionTarget.
The ViT-L variant further improves performance, demonstrating that enhancing gaze reasoning benefits both diagnostic subset evaluation and standard overall evaluation.

\begin{table}[!htbp]
\centering
\caption{
Overall comparison with state-of-the-art methods on GazeFollow and VideoAttentionTarget.
Results of prior methods are taken from GazeLLE~\citep{ryan2025gaze}, while our overall results are computed from the corresponding subset-level evaluations.
Learnable Params denotes the number of trainable parameters.
Input types are denoted as I: image, D: depth, E: eye, P: pose, and O: object.
With only RGB input and a small number of learnable parameters, our method achieves state-of-the-art localization performance on both benchmarks.
}
\label{tab:sota_all}
\footnotesize
\setlength{\tabcolsep}{3.5pt}
\renewcommand{\arraystretch}{1.08}
\resizebox{\textwidth}{!}{%
\begin{tabular}{@{}lclcccccc@{}}
\toprule
\multirow{2}{*}{\textbf{Method}}
& \multirow{2}{*}{\textbf{Learnable Params}}
& \multirow{2}{*}{\textbf{Input}}
& \multicolumn{3}{c}{\textbf{GazeFollow}}
& \multicolumn{3}{c}{\textbf{VideoAttentionTarget}} \\
\cmidrule(lr){4-6} \cmidrule(lr){7-9}
& &
& \textbf{AUC}\upb
& \textbf{Avg. L2}\downb
& \textbf{Min. L2}\downb
& \textbf{AUC}\upb
& \textbf{L2}\downb
& \textbf{AP}\textsubscript{\textbf{in/out}}\upb \\
\midrule

Recasens et al.~\citep{recasens2015they}
& 50M & I
& 0.878 & 0.190 & 0.113
& -- & -- & -- \\

Chong et al.~\citep{chong2018connecting}
& 51M & I
& 0.896 & 0.187 & 0.112
& 0.833 & 0.171 & 0.712 \\

Lian et al.~\citep{lian2018believe}
& 55M & I
& 0.906 & 0.145 & 0.081
& -- & -- & -- \\

Chong et al.~\citep{chong2020detecting}
& 61M & I
& 0.921 & 0.137 & 0.077
& 0.860 & 0.134 & 0.853 \\

Chen et al.~\citep{chen2021gaze}
& 50M & I
& 0.908 & 0.136 & 0.074
& -- & -- & -- \\

Fang et al.~\citep{fang2021dual}
& 68M & I+D+E
& 0.922 & 0.124 & 0.067
& 0.905 & 0.108 & 0.896 \\

Bao et al.~\citep{bao2022escnet}
& 29M & I+D+P
& 0.928 & 0.122 & --
& 0.885 & 0.120 & 0.869 \\

Jin et al.~\citep{jin2022depth}
& $>$52M & I+D+P
& 0.920 & 0.118 & 0.063
& 0.900 & 0.104 & 0.895 \\

Tonini et al.~\citep{tonini2023object}
& 92M & I+D
& 0.927 & 0.141 & --
& 0.862 & 0.125 & 0.742 \\

Hu et al.~\citep{hu2022gaze}
& $>$61M & I+D+O
& 0.923 & 0.128 & 0.069
& 0.880 & 0.118 & 0.881 \\

Gupta et al.~\citep{gupta2022modular}
& 35M & I+D+P
& 0.943 & 0.114 & 0.056
& 0.914 & 0.110 & 0.879 \\

Horanyi et al.~\citep{horanyi2023they}
& 46M & I+D
& 0.896 & 0.196 & 0.127
& 0.832 & 0.199 & 0.800 \\

Miao et al.~\citep{miao2023patch}
& 61M & I+D
& 0.934 & 0.123 & 0.065
& 0.917 & 0.109 & \textbf{0.908} \\

Tafasca et al.~\citep{tafasca2023childplay}
& $>$25M & I+D
& 0.939 & 0.122 & 0.062
& 0.914 & 0.109 & 0.834 \\

Tafasca et al.~\citep{tafasca2024sharingan}
& 105M & I
& 0.944 & 0.113 & 0.057
& -- & 0.107 & 0.891 \\

GazeLLE (ViT-B)~\citep{ryan2025gaze}
& 2.8M & I
& 0.956 & 0.104 & 0.045
& 0.933 & 0.107 & 0.897 \\

GazeLLE (ViT-L)~\citep{ryan2025gaze}
& 2.9M & I
& \underline{0.958} & 0.099 & \underline{0.041}
& \underline{0.937} & 0.103 & 0.903 \\

\midrule

\rowcolor{gray!15}
\textbf{Ours (ViT-B)}
& 3.7M & I
& \underline{0.958} & \underline{0.097} & \underline{0.041}
& 0.935 & \underline{0.101} & 0.899 \\

\rowcolor{gray!15}
\textbf{Ours (ViT-L)}
& 5.3M & I
& \textbf{0.960} & \textbf{0.090} & \textbf{0.037}
& \textbf{0.939} & \textbf{0.095} & \underline{0.907} \\

\bottomrule
\end{tabular}%
}
\end{table}

\subsection{Hyperparameter Sensitivity}
\label{app:hyper_sensitivity}

We provide a hyperparameter sensitivity study in Table~\ref{tab:hyper_sensitivity}.
We vary one hyperparameter at a time, including the out-of-cone loss weight $\lambda$, cone angle $\theta$, and the number of last adapted layers $L$ used for gaze-aware supervision.
The results show that our method is generally robust to these choices.
Our default setting, $\lambda=0.02$, $\theta=30^\circ$, and $L=4$, achieves the best performance on the inconsistent subset for most metrics while maintaining competitive performance on the consistent subset.

\begin{table}[!htbp]
\centering
\caption{
Hyperparameter sensitivity on GazeFollow.
We vary one hyperparameter at a time, including the loss weight $\lambda$, cone angle $\theta$, and the number of last adapted layers $L$ used for gaze-aware supervision, with the sharpness parameter fixed as $\alpha=40$.
A-L2/M-L2 and A-Ang/M-Ang denote average/minimum L2 distance and average/minimum angular error, respectively.
Bold highlights the metrics where our default choice is optimal within each hyperparameter group.
}
\label{tab:hyper_sensitivity}
\footnotesize
\setlength{\tabcolsep}{2.2pt}
\renewcommand{\arraystretch}{1.08}
\resizebox{\textwidth}{!}{%
\begin{tabular}{@{}l ccccc ccccc@{}}
\toprule
\multirow{2}{*}{\textbf{Setting}}
& \multicolumn{5}{c}{\textbf{Consistent}}
& \multicolumn{5}{c}{\textbf{Inconsistent}} \\
\cmidrule(lr){2-6} \cmidrule(lr){7-11}
& \textbf{AUC}\upb
& \textbf{A-L2}\downb
& \textbf{M-L2}\downb
& \textbf{A-Ang}\downb
& \textbf{M-Ang}\downb
& \textbf{AUC}\upb
& \textbf{A-L2}\downb
& \textbf{M-L2}\downb
& \textbf{A-Ang}\downb
& \textbf{M-Ang}\downb \\
\midrule

$\lambda=0.005$
& 0.9678 & \textbf{0.0777} & 0.0354 & 10.53 & 3.24
& 0.9311 & 0.1363 & 0.0571 & 19.03 & 5.60 \\

\rowcolor{gray!15}
$\lambda=0.02$
& 0.9696 & \textbf{0.0777} & \textbf{0.0348} & \textbf{10.50} & \textbf{3.19}
& \textbf{0.9340} & \textbf{0.1353} & \textbf{0.0547} & \textbf{18.00} & \textbf{4.85} \\

$\lambda=0.05$
& \textbf{0.9705} & 0.0795 & 0.0365 & 10.68 & 3.31
& 0.9325 & 0.1423 & 0.0585 & 19.61 & 6.13 \\

\midrule

$\theta=20^\circ$
& \textbf{0.9700} & \textbf{0.0756} & \textbf{0.0335} & \textbf{10.22} & \textbf{3.00}
& 0.9338 & 0.1383 & 0.0569 & 19.11 & 5.72 \\

\rowcolor{gray!15}
$\theta=30^\circ$
& 0.9696 & 0.0777 & 0.0348 & 10.50 & 3.19
& \textbf{0.9340} & \textbf{0.1353} & \textbf{0.0547} & \textbf{18.00} & \textbf{4.85} \\

$\theta=40^\circ$
& \textbf{0.9700} & 0.0772 & 0.0337 & 10.40 & 3.05
& 0.9334 & 0.1388 & 0.0573 & 18.48 & 5.29 \\

\midrule

$L=2$
& \textbf{0.9701} & 0.0802 & 0.0359 & 10.84 & 3.38
& \textbf{0.9343} & 0.1393 & 0.0559 & 19.16 & 5.81 \\

\rowcolor{gray!15}
$L=4$
& 0.9696 & 0.0777 & \textbf{0.0348} & 10.50 & 3.19
& 0.9340 & \textbf{0.1353} & \textbf{0.0547} & \textbf{18.00} & \textbf{4.85} \\

$L=8$
& 0.9690 & \textbf{0.0772} & 0.0350 & \textbf{10.49} & \textbf{3.15}
& 0.9319 & 0.1377 & 0.0565 & 18.61 & 5.30 \\

\bottomrule
\end{tabular}%
}
\end{table}

\subsection{Additional Analysis of Head Integration and Branch Design}
\label{app:head_branch_design}

We further analyze whether stronger architectural head integration is sufficient to improve gaze reasoning.
Following the design ablation protocol of GazeLLE~\citep{ryan2025gaze}, we compare different ways of incorporating the queried head information while keeping the decoder configuration consistent with their ablation setting.
Specifically, we consider early head integration before the VFM encoder, late head integration after scene feature extraction, and whether to use a separate head-and-scene branch or a single scene branch.

As shown in Table~\ref{tab:gazelle_design_ablation}, early head integration substantially degrades performance on both subsets, suggesting that injecting head information before the VFM encoder can disturb the pretrained visual representation.
Late integration performs much better, but adding a separate head branch brings only limited gains and still leaves a large gap on the inconsistent subset.
This supports our motivation that architectural head fusion alone is insufficient for robust gaze reasoning.

\begin{table}[!htbp]
\centering
\caption{
GazeLLE design variants on GazeFollow under our consistent/inconsistent evaluation.
Following the ablation setting of GazeLLE~\citep{ryan2025gaze}, we compare early vs. late head integration and single scene branch (S) vs. head-and-scene branches (H+S).
Early integration clearly hurts performance, while late H+S fusion still leaves a large gap on the inconsistent subset.
}
\label{tab:gazelle_design_ablation}
\scriptsize
\setlength{\tabcolsep}{1.8pt}
\renewcommand{\arraystretch}{1.05}
\resizebox{\textwidth}{!}{%
\begin{tabular}{@{}lcc ccccc ccccc@{}}
\toprule
\multirow{2}{*}{\textbf{Variant}}
& \multirow{2}{*}{\textbf{Head}}
& \multirow{2}{*}{\textbf{Br.}}
& \multicolumn{5}{c}{\textbf{Consistent}}
& \multicolumn{5}{c}{\textbf{Inconsistent}} \\
\cmidrule(lr){4-8} \cmidrule(lr){9-13}
& &
& \textbf{AUC}\upb
& \textbf{A-L2}\downb
& \textbf{M-L2}\downb
& \textbf{A-Ang}\downb
& \textbf{M-Ang}\downb
& \textbf{AUC}\upb
& \textbf{A-L2}\downb
& \textbf{M-L2}\downb
& \textbf{A-Ang}\downb
& \textbf{M-Ang}\downb \\
\midrule

GazeLLE
& Early & H+S
& 0.9213 & 0.1657 & 0.1079 & 18.21 & 9.30
& 0.8847 & 0.1905 & 0.1085 & \underline{23.82} & 9.97 \\

GazeLLE
& Late & H+S
& \textbf{0.9695} & \textbf{0.0875} & \underline{0.0400} & \textbf{11.54} & \textbf{3.74}
& \textbf{0.9286} & \textbf{0.1599} & \textbf{0.0743} & \textbf{22.15} & \textbf{8.14} \\

\rowcolor{gray!15}
GazeLLE
& Late & S
& \underline{0.9682} & \textbf{0.0875} & \textbf{0.0399} & \underline{11.99} & \underline{3.98}
& \underline{0.9219} & \underline{0.1670} & \underline{0.0783} & 24.05 & \underline{9.65} \\

\bottomrule
\end{tabular}%
}
\end{table}

\begin{samepage}
We also examine adding a separate head-and-scene branch to our method.
As shown in Table~\ref{tab:ours_head_scene_branch}, this design slightly improves a few metrics on the consistent subset, but degrades the inconsistent subset.
For example, Avg.~L2 increases from 0.1353 to 0.1421 and Min.~Ang. from 4.85 to 6.35.
Therefore, our final model keeps the simpler single-branch design and uses HCLoRA to inject head-conditioned adaptation inside the VFM.
\par
\end{samepage}

\begin{table}[!htbp]
\centering
\caption{
Effect of adding a separate head-and-scene branch to our method on GazeFollow.
The H+S branch slightly improves consistent-subset metrics but degrades the inconsistent subset.
}
\label{tab:ours_head_scene_branch}
\tiny
\setlength{\tabcolsep}{1.4pt}
\renewcommand{\arraystretch}{0.88}
\resizebox{1.0\textwidth}{!}{%
\begin{tabular}{@{}l ccccc ccccc@{}}
\toprule
\multirow{2}{*}{\textbf{Variant}}
& \multicolumn{5}{c}{\textbf{Consistent}}
& \multicolumn{5}{c}{\textbf{Inconsistent}} \\
\cmidrule(lr){2-6} \cmidrule(lr){7-11}
& \textbf{AUC}\upb
& \textbf{A-L2}\downb
& \textbf{M-L2}\downb
& \textbf{A-Ang}\downb
& \textbf{M-Ang}\downb
& \textbf{AUC}\upb
& \textbf{A-L2}\downb
& \textbf{M-L2}\downb
& \textbf{A-Ang}\downb
& \textbf{M-Ang}\downb \\
\midrule

\rowcolor{gray!15}
Ours
& 0.9696 & \textbf{0.0777} & 0.0348 & \textbf{10.50} & 3.19
& \textbf{0.9340} & \textbf{0.1353} & \textbf{0.0547} & \textbf{18.00} & \textbf{4.85} \\

\rowcolor{gray!15}
Ours + H+S
& \textbf{0.9704} & 0.0790 & \textbf{0.0345} & 10.63 & \textbf{3.04}
& 0.9310 & 0.1421 & 0.0610 & 19.70 & 6.35 \\

\bottomrule
\end{tabular}%
}
\end{table}

\subsection{Limitations in Challenging Scenarios}
\label{app:failure_limitations}

\begin{figure}[!htbp]
    \centering
    \includegraphics[width=\linewidth]{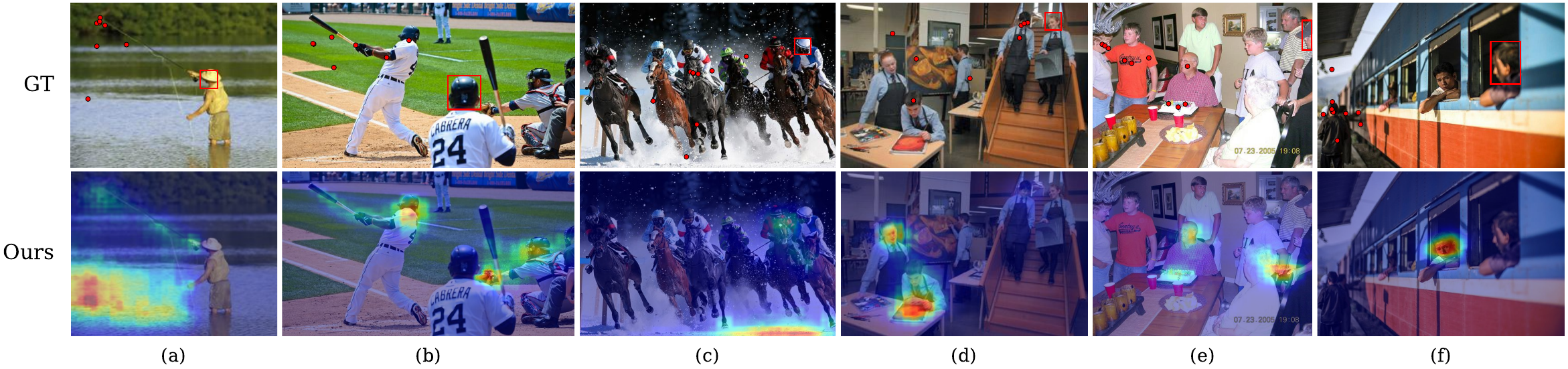}
    \caption{Challenging cases for our method on GazeFollow. The top row shows the queried head boxes and ground-truth gaze annotations in red; the bottom row shows our predicted heatmaps. Weak or ambiguous head cues can limit gaze localization in these cases.}
    \label{fig:ours_failure_cases}
\end{figure}

By incorporating head-conditioned adaptation, our method achieves strong overall gaze localization performance, with particularly strong gains on the inconsistent subsets.
Nevertheless, as illustrated in Figure~\ref{fig:ours_failure_cases}, it can still struggle when the queried head is small, blurred, partially occluded, or viewed from behind, making gaze direction difficult to infer.
These challenging scenarios leave room for further improvement in exploiting limited head cues.

\section{Broader Impacts}

Gaze following is important for understanding human attention and behavior, with potential benefits for human-robot interaction, assistive systems, educational technologies, and human-centered scene understanding.

At the same time, gaze following is privacy-sensitive. If deployed without safeguards, such models could be misused for surveillance, behavioral profiling, or unintended inference of personal attention patterns, especially in public spaces, workplaces, and schools where explicit consent may be absent. Gaze behavior may also vary across individuals, contexts, cultures, and visual conditions, so benchmark-trained models may not generalize equally well across populations or environments.

Our work uses existing benchmark datasets and does not collect new personal data. Real-world deployment should require clear user consent, data protection measures, and careful consideration of privacy, fairness, and potential misuse.

\end{document}